\documentclass{article}

\usepackage{arxiv}

\usepackage[utf8]{inputenc} 
\usepackage[T1]{fontenc}    
\usepackage{amsmath}
\usepackage{hyperref}       
\usepackage{url}            
\usepackage{booktabs}       
\usepackage{tabularx}
\usepackage{amsfonts}       
\usepackage{nicefrac}       
\usepackage{microtype}      
\usepackage{graphicx}
\usepackage{placeins}
\usepackage{natbib}
\usepackage{doi}

 \title{RHyVE: Competence-Aware Verification and Phase-Aware Deployment for LLM-Generated Reward Hypotheses}
 
 \date{}
 
\author{
	\href{https://orcid.org/0009-0001-9198-4783}{\includegraphics[scale=0.06]{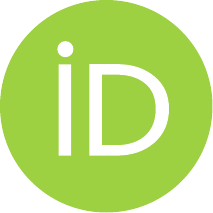}\hspace{1mm}Feiyu Wu}
 	\hspace{0.75em}
	\href{https://orcid.org/0009-0001-9202-7088}{\includegraphics[scale=0.06]{orcid.pdf}\hspace{1mm}Xu Zheng}
 	\hspace{0.75em}
	\href{https://orcid.org/0009-0009-2355-1191}{\includegraphics[scale=0.06]{orcid.pdf}\hspace{1mm}Zhuocheng Wang}
 	\hspace{0.75em}
	\href{https://orcid.org/0009-0001-0732-8223}{\includegraphics[scale=0.06]{orcid.pdf}\hspace{1mm}Yi ming Dai}
 	\hspace{0.75em}
	\href{https://orcid.org/0000-0001-8310-7169}{\includegraphics[scale=0.06]{orcid.pdf}\hspace{1mm}Hui Li}\thanks{Corresponding author: lihui@mail.xidian.edu.cn} \\
	School of Cyber Engineering, Xidian University \\
	\texttt{sn0wm1ans@gmail.com, zhengxu200477@gmail.com, smilencet1@gmail.com} \\
	\texttt{d18797323123@qq.com, lihui@mail.xidian.edu.cn}}
 
 \renewcommand{\shorttitle}{RHyVE}

\hypersetup{
pdftitle={RHyVE: Competence-Aware Verification and Phase-Aware Deployment for LLM-Generated Reward Hypotheses},
pdfsubject={Machine Learning},
pdfauthor={Feiyu Wu, Xu Zheng, Zhuocheng Wang, Yi ming Dai, Hui Li},
pdfkeywords={reinforcement learning, reward hypotheses, LLM-generated rewards, verification, phase-aware deployment},
}

\begin{document}
\maketitle
\begin{abstract}
Large language models (LLMs) make reward design in reinforcement learning substantially more scalable, but generated rewards are not automatically reliable training objectives. Existing work has focused primarily on generating, evolving, or selecting reward candidates, while paying less attention to when such candidates can be verified and deployed during policy optimization. We study this deployment-time problem by treating generated rewards as reward hypotheses whose utility depends on the competence of the current policy and the phase of training. We propose \textsc{RHyVE}, a competence-aware verification and phase-aware deployment protocol that compares small sets of reward hypotheses from shared policy checkpoints using short-horizon fork verification. Our experiments show that reward rankings are unreliable at low competence but become informative after task-dependent thresholds. On a sparse manipulation task, phase-aware deployment improves peak and retained performance under a locked protocol. Updated LLM-generated reward-candidate experiments show candidate-family-dependent behavior: generated pools can exhibit phase-dependent winner changes, but no fixed warm-up schedule is universally optimal. Held-out schedule selection, conservative selector baselines, compute-matched controls, and scale controls further show that \textsc{RHyVE} is best understood as a verification-informed deployment protocol rather than a universal scheduler. Dense and all-failure boundary experiments delimit the scope of the method. Together, these results suggest that reward generation and reward deployment should be studied as coupled problems: generated rewards must be verified and deployed under changing policy competence.
\end{abstract}

\section{Introduction}

Reward design remains a central bottleneck in reinforcement learning: good shaping rewards can accelerate learning, while misspecified rewards can induce proxy optimization, reward hacking, and poor generalization~\cite{ng1999policy,wiewiora2003potential,devlin2012dynamic,amodei2016concrete,everitt2017reinforcement,everitt2019reward,skalse2022defining,pan2024feedback}. A broad literature studies how to obtain rewards, including inverse RL, preference-based RL, RLHF, and structured reward specification~\cite{ng2000algorithms,abbeel2004apprenticeship,ziebart2008maximum,finn2016guided,christiano2017deep,ibarz2018reward,stiennon2020learning,ouyang2022training,bai2022constitutional,toro2018using,camacho2019ltl,toro2022reward,bukharin2025deep}. Yet once a candidate reward is available, an equally practical question remains: when is it reliable enough to drive optimization?

This question is especially important for LLM-based reward design. Recent systems synthesize reward code, dense shaping functions, and online reward selectors from language~\cite{ma2024eureka,xie2024text2reward,li2024automc,hazra2025revolve,zhang2025orso,li2025rstar,rocamonde2024vision}. These methods can generate plausible candidates, but usually treat reward quality as a property of the candidate alone: generate a set of rewards, evaluate or refine them, and deploy the selected one. Our central observation is that this view is incomplete. The apparent quality of a generated reward can depend strongly on the competence of the current policy and on the phase of training in which the reward is used.

We therefore argue that generated rewards should be treated as \emph{reward hypotheses} rather than immediately reliable objectives. A reward that is useful for late-stage optimization may be too sparse, too sharp, or too poorly scaled to bootstrap learning from a weak policy. Conversely, a dense reward that exposes coarse progress signals early in training may later become misaligned with the task objective. This view connects reward deployment to curricula and adaptive objectives~\cite{bengio2009curriculum,florensa2017reverse,graves2017automated,matiisen2017teacher,jaderberg2017population,wang2020comprehensive,xu2018meta,zheng2018learning}, but sharpens the deployment question in the reward-design setting: the issue is not only \emph{which} reward is best, but also \emph{when} the learner is competent enough to verify that fact.

A second deployment challenge arises when the training objective changes. Switching rewards mid-training can create non-stationary targets for the actor and critic, produce reward-scale shocks, and destabilize value estimation. Potential-based shaping and trust-region methods motivate caution, but they do not guarantee that an immature learned critic can safely support an online reward switch~\cite{schulman2015trust,achiam2017constrained,ng1999policy,wiewiora2003potential,devlin2012dynamic}. Reward deployment therefore requires attention both to verification timing and to switching stability.

We therefore propose \textsc{RHyVE}, a competence-aware verification and phase-aware deployment protocol for small sets of reward hypotheses. \textsc{RHyVE} compares candidates from shared policy checkpoints using short-horizon fork verification, records when winner identity becomes reliable, and uses the resulting phase profile to decide whether a single reward suffices or whether a two-stage schedule is warranted. The method is deliberately local in scope: it is not a new reward generator and not a universal adaptive scheduler, but a protocol for deciding when already available candidates are trustworthy enough to compare and deploy. In this sense, \textsc{RHyVE} addresses the missing middle between reward generation and reward commitment.

This paper makes the following contributions:
\begin{itemize}
    \item We formulate generated rewards as hypotheses whose reliability depends on policy competence and training phase, separating reward generation from reward verification and deployment.
    \item We introduce shared-checkpoint fork verification and phase-profile construction for small candidate sets, together with deployment rules that choose a single reward, a two-stage schedule, or conservative fallback depending on verification reliability.
    \item We show on the locked \textsc{FrankaCabinet} protocol that phase-aware deployment improves both peak and recomputed terminal performance in a sparse phase-sensitive manipulation regime.
    \item We extend the evidence to LLM-generated reward-candidate families under a reduced 6$\times$3090 protocol, showing candidate-family-dependent deployment behavior rather than universal warm-up optimality.
    \item We provide controls and scope evidence, including held-out schedule selection, conservative online selector baselines, compute-matched direct training, reward-scale controls, dense-oracle boundary behavior, candidate-pool stress tests, and an all-failure extra-task pilot.
\end{itemize}
\section{Related Work}

Recent work shows that large language models and foundation models can automate substantial parts of reward engineering. Eureka, Text2Reward, Auto MC-Reward, REvolve, and R* synthesize or refine executable reward programs and shaping functions; ORSO is the closest operational prior because it studies online reward selection during training; and pretrained vision-language models can act as zero-shot reward models. These works mainly study how to generate, refine, search over, or reactively select reward candidates. \textsc{RHyVE} studies a complementary deployment-time question: when does a small candidate set become reliable to compare from the current learner state, and should the resulting evidence lead to a single reward, a phase-aware schedule, or conservative fallback? Our conservative selector baselines are included to test reactive alternatives, but we do not claim to reproduce or dominate the full ORSO system.

Learned and structured reward methods reduce manual reward engineering by inferring or constraining reward signals from demonstrations, preferences, human feedback, or formal task structure~\cite{ng2000algorithms,abbeel2004apprenticeship,ziebart2008maximum,finn2016guided,christiano2017deep,ibarz2018reward,stiennon2020learning,ouyang2022training,bai2022constitutional,hadfieldmenell2017inverse,toro2018using,camacho2019ltl,toro2022reward,bukharin2025deep}. Curriculum, teacher-student, population-based, and meta-gradient methods likewise exploit the fact that useful supervision can change during learning~\cite{bengio2009curriculum,florensa2017reverse,graves2017automated,matiisen2017teacher,jaderberg2017population,wang2020comprehensive,xu2018meta,zheng2018learning}. Our focus is narrower: given already available rewards, we ask when the current learner is competent enough for local comparison, when reward utility changes across phases, and when a fixed phase-aware deployment rule is warranted.

Reward shaping and safe optimization are directly relevant once the objective changes. Potential-based shaping gives policy-invariance guarantees under suitable assumptions~\cite{ng1999policy,wiewiora2003potential,devlin2012dynamic,grzes2010online}, while trust-region, constrained, and safety-oriented perspectives stress that changing objectives can induce instability, proxy optimization, or reward tampering~\cite{schulman2015trust,achiam2017constrained,amodei2016concrete,everitt2017reinforcement,everitt2019reward,skalse2022defining,pan2024feedback}. \textsc{RHyVE} does not propose a universal switch operator; instead, it separates competence-aware verification from the trade-offs of how a reward change is executed, which is why our experiments treat hard switching, PBRS, and critic reset as conditional mechanisms rather than interchangeable defaults.
\section{Problem Setup and Method}
\label{sec:method}

In \textsc{RHyVE}, we study the deployment-time problem that arises after an upstream reward-design procedure, such as an LLM-based reward generator, has produced a small set of candidate rewards. Our goal is not to generate new reward code, but to decide when candidate rewards are reliable to compare, when they should be deployed, and how objective changes should be introduced without destabilizing training. The central distinction is that candidate rewards are treated as hypotheses whose usefulness may depend on learner competence rather than as immediately trustworthy objectives.

The method is intentionally modest in scope. \textsc{RHyVE} is designed for the practically common regime in which a small number of structured candidate rewards is already available and the learner's competence changes over training.

\subsection{Reward Hypotheses}
\label{sec:reward-hypotheses}

We consider an episodic Markov decision process
\[
\mathcal{M} = (\mathcal{S}, \mathcal{A}, P, \rho_0, \gamma),
\]
where $\mathcal{S}$ and $\mathcal{A}$ are the state and action spaces, $P$ is the transition kernel, $\rho_0$ is the initial-state distribution, and $\gamma \in (0,1)$ is the discount factor. Instead of assuming a single fixed reward, we are given a small candidate set
\[
\mathcal{H} = \{h_1,\ldots,h_K\},
\]
where each hypothesis $h_k$ induces a reward function
\[
r_{h_k}: \mathcal{S}\times\mathcal{A}\times\mathcal{S}\rightarrow \mathbb{R}.
\]
We use the term \emph{reward hypothesis} deliberately. A generated reward is not assumed to be immediately trustworthy, globally optimal, or useful at every training phase. Rather, it is a candidate training signal whose value may depend on the competence of the current policy.

Let
\[
z_t = (\pi_t, V_t, \Omega_t)
\]
denote a training checkpoint at optimization step $t$, consisting of the current policy $\pi_t$, critic $V_t$, and optimizer or training state $\Omega_t$. Associated with each checkpoint is a task-dependent competence proxy
\[
c_t = c(z_t),
\]
such as success rate, task progress, or another downstream metric. We do not assume a universal competence scale across tasks; competence is used only to determine whether the current learner is in a regime where reward comparisons are informative.

\begin{figure}[t]
    \centering
    \includegraphics[width=\textwidth]{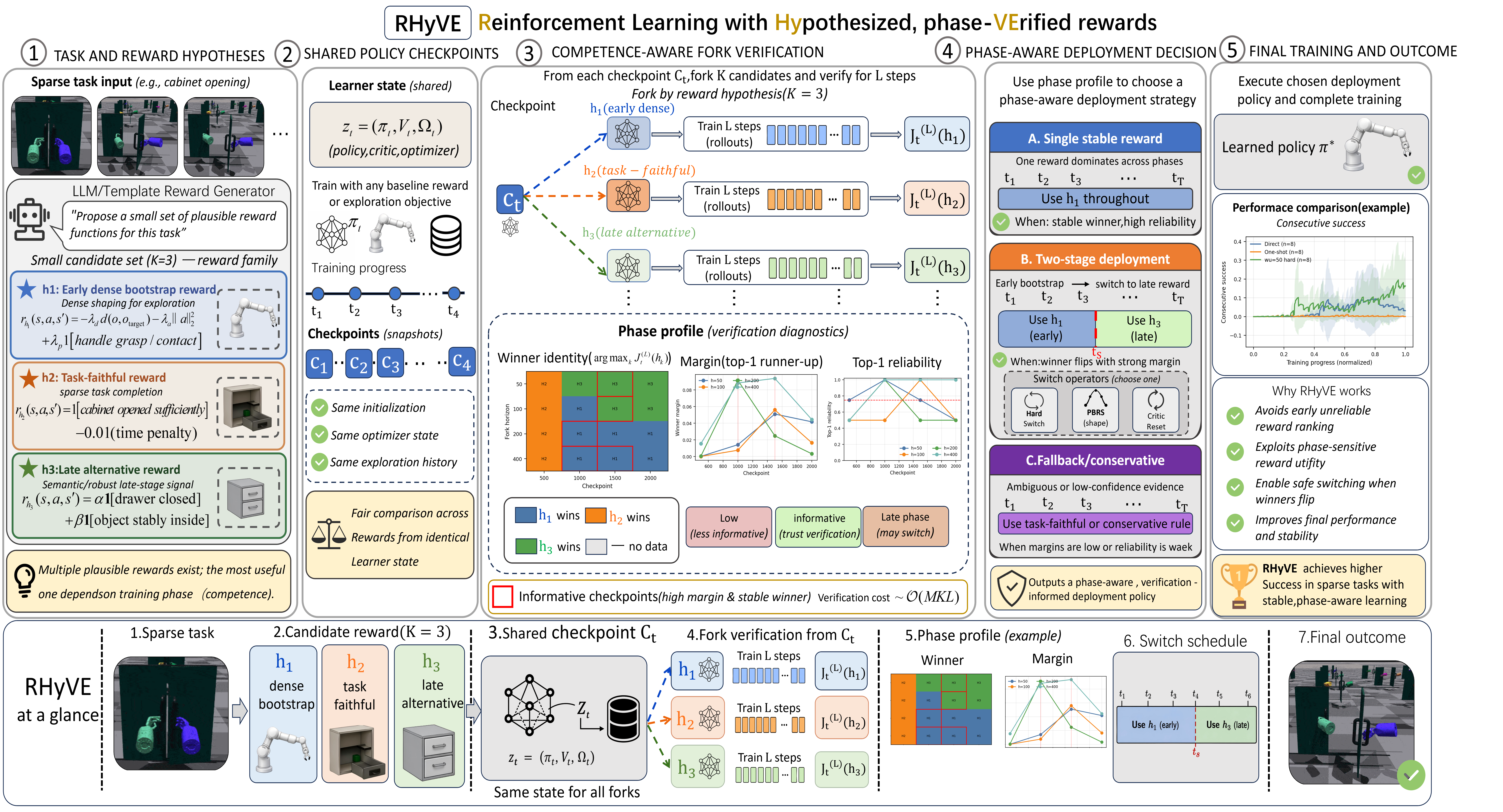}
    \caption{Overview of \textsc{RHyVE}. Reward candidates are treated as hypotheses, compared from shared checkpoints using fork verification, and deployed as a single reward, two-stage schedule, or conservative fallback depending on the phase profile.}
    \label{fig:fig1}
\end{figure}

\subsection{Shared-Checkpoint Fork Verification}
\label{sec:fork-verification}

The central operation in \textsc{RHyVE} is shared-checkpoint fork verification. At a checkpoint $z_t$, each reward hypothesis $h_k$ is evaluated by cloning the same learner state, continuing training for a short fork horizon $L$ under reward $r_{h_k}$, and evaluating the resulting forked policy with a common downstream metric:
\[
J_t^{(L)}(h_k)
=
\operatorname{Eval}\!\left(
    \operatorname{Train}(z_t, r_{h_k}, L)
\right).
\]
The local winner at checkpoint $t$ is
\[
\hat{h}_t
=
\arg\max_{h_k\in\mathcal{H}} J_t^{(L)}(h_k),
\]
and its margin over the runner-up is
\[
m_t
=
J_t^{(L)}(\hat{h}_t)
-
\max_{h_j\neq \hat{h}_t} J_t^{(L)}(h_j).
\]
Shared checkpoints avoid confounding reward quality with independent initialization, exploration history, and optimizer state. The fork horizon $L$ is intentionally short: \textsc{RHyVE} estimates which reward is locally useful for the current learner rather than fully optimizing every candidate. This keeps verification aligned with the practical question of whether a reward should be deployed from the current policy state.

If $M$ checkpoints are probed, $K$ hypotheses are compared, and each fork runs for $L$ update steps, the additional verification cost scales as
\[
O(MKL).
\]
Thus \textsc{RHyVE} is designed for small candidate sets, sparse checkpoint probes, and short local verification windows.

\subsection{Verification-Informed Phase Profiles}
\label{sec:phase-profile}

Fork verification produces a phase profile over training:
\[
\mathcal{P}
=
\left\{
(t, c_t, \hat{h}_t, m_t, \{J_t^{(L)}(h_k)\}_{k=1}^{K})
:\, t\in\mathcal{T}
\right\},
\]
where $\mathcal{T}=\{t_1,\ldots,t_M\}$ is a sparse set of probed checkpoints. This profile records how apparent reward utility changes as the policy becomes more competent and provides a compact diagnostic of whether the candidate family exhibits a stable phase structure.

Early checkpoints may be uninformative because a weak policy has not yet reached states where reward differences matter. We therefore treat a checkpoint as verification-informative only when the comparison shows sufficient separation and stability, assessed using
\[
m_t,\qquad
\operatorname{Agree}_t,\qquad
\operatorname{Ent}_t,
\]
where $m_t$ is the winner margin, $\operatorname{Agree}_t$ is repeated-fork winner agreement, and $\operatorname{Ent}_t$ is winner entropy across repeated forks or nearby horizons. In practice, these diagnostics are used to distinguish true phase structure from noisy local rank fluctuations. If winner identity is unstable or margins are negligible, \textsc{RHyVE} abstains from aggressive reward commitment; if a winner transition stabilizes, the profile suggests phase-dependent reward utility.

\subsection{Phase-Aware Deployment}
\label{sec:phase-deployment}

Given a phase profile $\mathcal{P}$, \textsc{RHyVE} deploys a single reward when one hypothesis is stable from the first informative checkpoint onward, a two-stage schedule when an early winner is later overtaken by a stable later winner, and a conservative no-switch rule when the profile remains ambiguous. The two-stage case uses
\[
r_t =
\begin{cases}
r_{h^{(1)}} & t < t_s,\\
r_{h^{(2)}} & t \ge t_s,
\end{cases}
\]
where $t_s$ is the first stable later-phase switch point. In practice, we prefer deployment rules that are selected from phase profiles and then evaluated on held-out seeds, rather than rules chosen after inspecting test performance. When the phase profile is stable, \textsc{RHyVE} may deploy a two-stage schedule such as a fixed warm-up followed by a task-faithful reward. When a single candidate is stable from the first informative checkpoint onward, \textsc{RHyVE} deploys that single reward. When margins, agreement, or winner identity remain ambiguous, \textsc{RHyVE} falls back to a conservative rule rather than forcing a switch. This framing is important: \textsc{RHyVE} is not a universal scheduler, and fixed warm-up is not assumed to be optimal for every candidate family.

\subsection{Switch Operators}
\label{sec:switch-operators}

Once a later-phase reward has been identified, the remaining question is how to execute the objective change. Hard switching directly replaces the early reward with the later reward,
\[
r_t =
\begin{cases}
r_{h^{(1)}} & t < t_s,\\
r_{h^{(2)}} & t \ge t_s.
\end{cases}
\]
and serves as the practical default in our experiments because it introduces no additional learned shaping terms. A smoothing alternative uses potential-based shaping~\cite{ng1999policy,wiewiora2003potential,devlin2012dynamic}: given a source critic $V^{\mathrm{old}}_{t_s}$ trained before the switch, we define
\[
\tilde{r}_{h^{(2)}}(s,a,s')
=
r_{h^{(2)}}(s,a,s')
+
\gamma \Phi(s') - \Phi(s),
\qquad
\Phi(s)=V^{\mathrm{old}}_{t_s}(s).
\]
This operator is useful only when the source critic is sufficiently informative; otherwise the shaping term can inject stale or noisy value estimates. A third operator keeps the actor but reinitializes the critic at the switch point,
\[
(\pi^{\mathrm{new}}_{t_s}, V^{\mathrm{new}}_{t_s})
=
(\pi^{\mathrm{old}}_{t_s}, V_{\mathrm{reset}}).
\]

These operators are therefore analyzed as conditional mechanisms. The experiments treat hard switching, PBRS, and critic reset as stability--retention trade-offs rather than as interchangeable safe-switching guarantees.
\section{Experiments}
\label{sec:experiments}

We evaluate \textsc{RHyVE} as a deployment-time protocol for small sets of reward hypotheses. The experiments are designed to answer four questions: when fork verification becomes informative, whether phase-aware deployment helps in a sparse phase-sensitive regime, whether the phenomenon appears for LLM-generated reward candidates, and which controls delimit the scope of the method.

\subsection{Experimental Setup}

The main task is \textsc{FrankaCabinet}, a sparse and phase-sensitive manipulation task in Isaac Gym. Unless otherwise stated, each task uses a small candidate set with $K=3$ reward hypotheses and PPO-style actor--critic optimization. The structured \textsc{FrankaCabinet} family contains an early dense bootstrap reward, a later task-faithful oracle-like reward, and a later alternative reward. Fork verification compares candidates from shared policy checkpoints over short horizons and scores the resulting forked policies with a common downstream metric.

We report peak success (\texttt{consec\_max}) together with recomputed final success and tail metrics when available. This distinction is important because a method can achieve a transient success spike and later collapse. Main-text tables report seed counts; evidence status is made explicit in the appendix taxonomy and appendix tables. Locked structured results support the main sparse-task claim; reduced 6$\times$3090 LLM results support candidate-family-dependent deployment behavior; selector, compute-matched, scale, and extra-task experiments are used as controls or scope evidence.

\subsection{Competence-Limited Reward Verification}

\begin{figure}[t]
    \centering
    \includegraphics[width=0.92\textwidth]{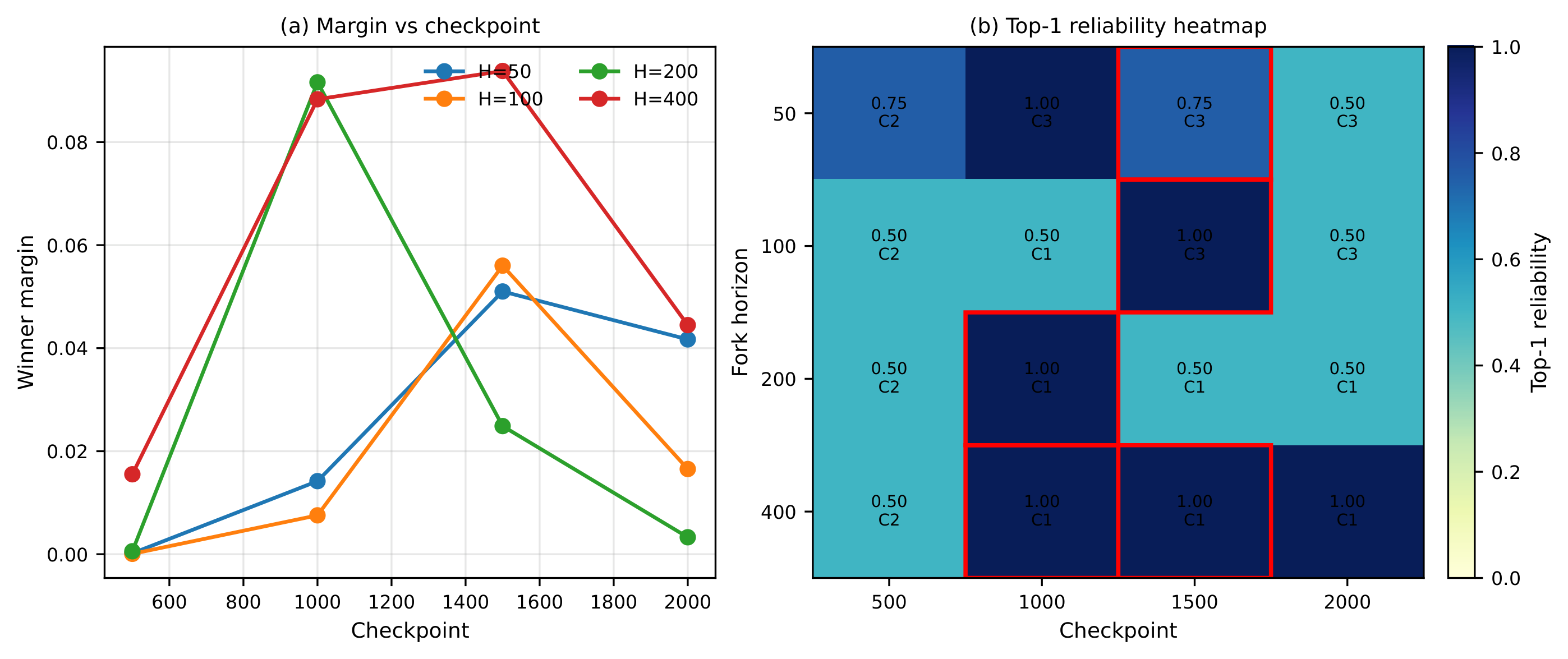}
    \caption{Competence-limited fork verification. Reward rankings are unstable at low competence and become informative only after task-dependent checkpoints and sufficient fork horizon.}
    \label{fig:fig2}
\end{figure}

Figure~\ref{fig:fig2} shows that fork-based reward rankings are competence-limited. On \textsc{FrankaCabinet}, early checkpoints are unstable or favor the dense bootstrap reward, while later checkpoints reveal a stable later-phase winner. The first informative checkpoint also depends on fork horizon. This motivates phase-aware deployment rather than one-shot reward commitment.

Two details are important. First, an early checkpoint can produce a modal winner without being verification-informative, because the winner margin or repeated-fork agreement may still be weak. Second, longer fork horizons can expose later-phase utility earlier than very short forks, but at higher verification cost. Thus reward verification is itself a competence-dependent diagnostic rather than a static evaluation of the candidate set.

\subsection{Phase-Aware Deployment on \textsc{FrankaCabinet}}

\begin{figure}[t]
    \centering
    \includegraphics[width=0.92\textwidth]{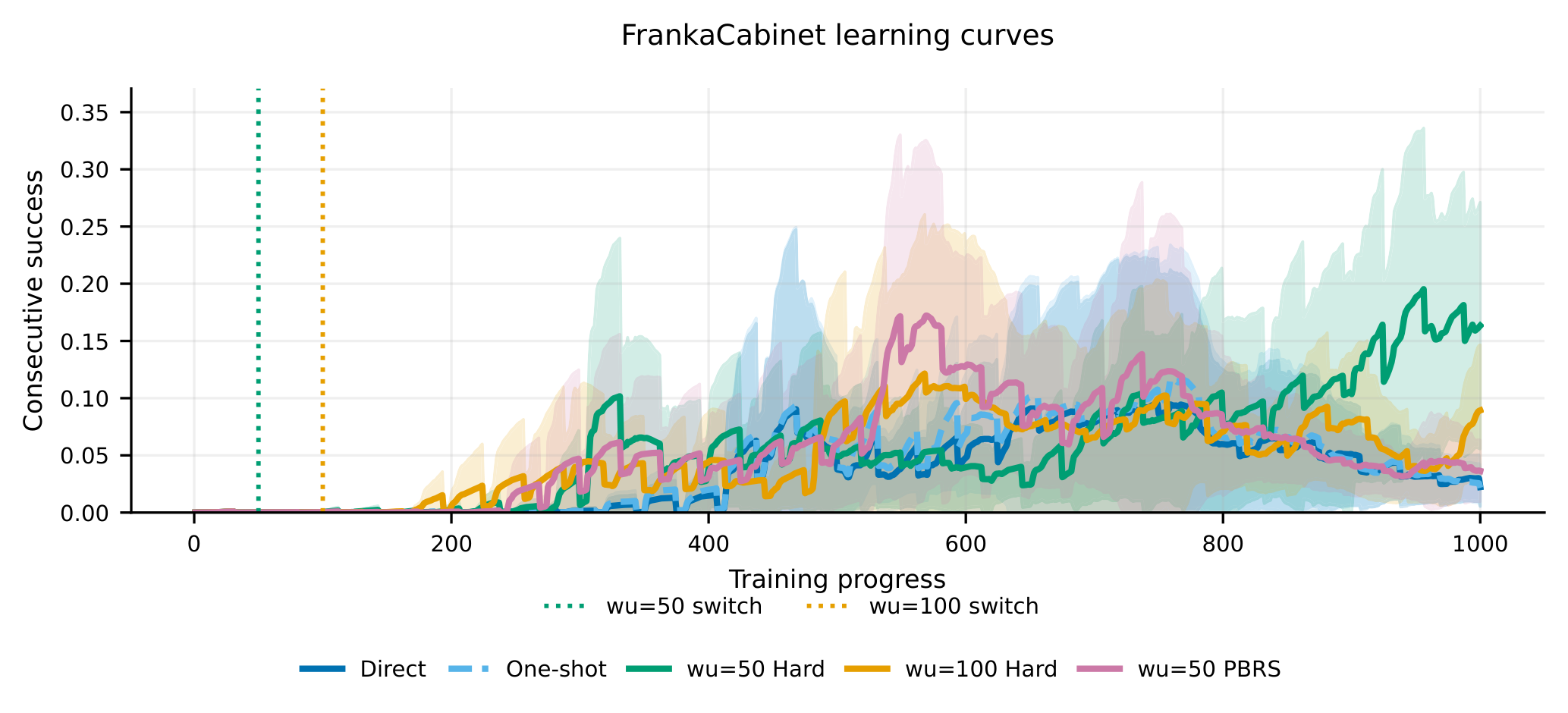}
    \caption{Locked FrankaCabinet learning curves. The phase-aware \texttt{wu=50} hard-switch schedule improves both peak and retained behavior relative to direct and one-shot deployment under the structured reward family.}
    \label{fig:fig3}
\end{figure}

Table~\ref{tab:table1} reports the locked \textsc{FrankaCabinet} structured comparison. The fixed \texttt{wu=50} hard-switch schedule achieves the strongest mean peak performance and the strongest recomputed final performance among the main locked methods. Figure~\ref{fig:fig3} shows the corresponding learning curves.

The result should be interpreted as a sparse-regime deployment finding rather than a universal scheduler claim. Peak paired tests remain affected by high seed variance, so we avoid claiming broad benchmark dominance. The practical conclusion is narrower: in this sparse phase-sensitive candidate family, a short competence-informed warm-up is a strong deployment default, and its gain is visible in both peak and retained behavior.

\begin{table}[t]
\centering
\caption{Locked FrankaCabinet structured-reward results. Values are means over fixed seeds. Final and tail metrics are recomputed from learning curves when available.}
\label{tab:table1}
\resizebox{\columnwidth}{!}{\begin{tabular}{lccccc}
\toprule
Method & $n$ & Max / \texttt{consec\_max} $\uparrow$ & Recomputed final $\uparrow$ & Last-5 mean $\uparrow$ & Tail AUC $\uparrow$ \\
\midrule
Direct oracle (\texttt{wu=0}) & 8 & 0.215 & 0.022 & 0.028 & 0.034 \\
One-shot oracle & 8 & 0.228 & 0.022 & 0.025 & 0.034 \\
Fixed \texttt{wu=50} hard switch & 8 & 0.389 & 0.163 & 0.161 & 0.157 \\
Fixed \texttt{wu=100} hard switch & 8 & 0.233 & 0.089 & 0.087 & 0.056 \\
\texttt{wu=50} + $V_{\mathrm{old}}$ PBRS & 8 & 0.314 & 0.036 & 0.037 & 0.040 \\
\bottomrule
\end{tabular}
}
\end{table}

\subsection{LLM-Generated Reward Candidate Families}

Table~\ref{tab:table2} reports the updated \textsc{FrankaCabinet} LLM-generated candidate results under the reduced 6$\times$3090 protocol. These results replace the older pilot-style source-generalization table. The updated rows show that generated candidate pools can exhibit phase-dependent deployment behavior, but the best deployable rule is candidate-family-dependent.

The lightly curated and minimally filtered LLM families should not be read as evidence that a fixed \texttt{wu=50} schedule is universally optimal. Instead, they support the central diagnostic claim: generated rewards are hypotheses whose usefulness depends on policy competence, candidate source, and deployment rule. Oracle-style adaptive rows, when shown, are separated from deployable methods and are not used as main method claims.

\begin{table}[t]
\centering
\caption{FrankaCabinet LLM-generated reward-candidate results under the reduced 6x3090 protocol. The best deployment rule varies across candidate families, supporting a verification-informed rather than universal-schedule interpretation. Oracle rows are non-deployable references and are not guaranteed to upper-bound deployable rows when seed sets or comparison protocols differ.}
\label{tab:table2}
\resizebox{\columnwidth}{!}{\begin{tabular}{llcccc}
\toprule
Candidate source & Method & $n$ & Max / \texttt{consec\_max} $\uparrow$ & Flip rate & Status \\
\midrule
\texttt{llm\_lightly\_curated} & Direct & 8 & 0.145 & 0.875 & deployable \\
\texttt{llm\_lightly\_curated} & \texttt{wu=50} hard & 8 & 0.207 & 0.875 & deployable \\
\texttt{llm\_lightly\_curated} & Held-out selected & 5 & 0.298 & -- & deployable \\
\cmidrule(lr){2-6}
\texttt{llm\_lightly\_curated} & \textit{Best adaptive (oracle reference)} & 3 & 0.580 & 0.875 & oracle \\
\midrule
\texttt{llm\_minimally\_filtered} & Direct & 8 & 0.211 & 0.500 & deployable \\
\texttt{llm\_minimally\_filtered} & \texttt{wu=50} hard & 8 & 0.639 & 0.500 & deployable \\
\texttt{llm\_minimally\_filtered} & Held-out selected & 5 & 0.800 & -- & deployable \\
\cmidrule(lr){2-6}
\texttt{llm\_minimally\_filtered} & \textit{Best adaptive (oracle reference)} & 2 & 0.369 & 0.500 & oracle \\
\bottomrule
\end{tabular}
}
\end{table}

\subsection{Held-Out Schedule Selection}

Table~\ref{tab:table3} audits schedule selection using dev seeds and held-out report seeds. This is included to reduce the risk that a fixed switch time is interpreted as post-hoc test-set tuning. The schedule selected from dev-seed phase profiles is frozen before evaluation on held-out seeds. If the phase profile is ambiguous, \textsc{RHyVE} uses a conservative fallback rather than forcing an aggressive switch.

This held-out audit is not meant to prove that a single switch time is universal. Its role is to show that deployment choices can be made from verification evidence without inspecting held-out performance, and that the resulting behavior can be reported separately from oracle upper bounds.

\begin{table}[t]
\centering
\small
\caption{Held-out schedule-selection audit. Schedules are selected using dev seeds and evaluated on held-out report seeds; only deployable held-out rows are shown in the main text.}
\label{tab:table3}
\begin{tabularx}{\textwidth}{@{}>{\raggedright\arraybackslash}p{0.34\textwidth}>{\raggedright\arraybackslash}X>{\centering\arraybackslash}p{0.08\textwidth}>{\centering\arraybackslash}p{0.16\textwidth}@{}}
\toprule
Candidate source & Deployable rule & $n$ & Max / \texttt{consec\_max} $\uparrow$ \\
\midrule
\texttt{structured\_handcrafted\_K3} & \texttt{fixed\_wu50\_hard} $\rightarrow$ \texttt{fixed\_wu50\_hard} & 5 & 0.394 \\
\texttt{llm\_lightly\_curated} & \texttt{direct\_oracle\_or\_direct\_best} $\rightarrow$ \texttt{direct\_oracle\_or\_direct\_best} & 5 & 0.125 \\
\texttt{llm\_minimally\_filtered} & \texttt{fixed\_wu50\_hard} $\rightarrow$ \texttt{fixed\_wu50\_hard} & 5 & 0.800 \\
\bottomrule
\end{tabularx}

\end{table}

\subsection{Robustness and Fairness Controls}

Table~\ref{tab:table4} summarizes three important controls. First, compute-matched direct training tests whether the advantage of phase-aware deployment is simply due to the extra fork-verification budget. The compute-matched direct baseline narrows the gap, showing that verification overhead is a real fairness consideration, but it does not by itself establish that direct deployment is preferable. Second, conservative online selector baselines test whether reactive reward selection can replace stable phase-aware deployment in this sparse small-candidate regime. These selectors underperform the stable schedule in the updated support comparison, and their failure modes are analyzed in the appendix. Third, scale controls show that reward scale is an important confound, but scale matching does not fully explain the phase-aware gain.

Together, these controls support a cautious interpretation: \textsc{RHyVE} is useful because it separates verification, deployment, and switching stability. It should not be interpreted as a claim that compute is irrelevant, that reactive selectors never work, or that scale mismatch is unimportant.

\begin{table}[t]
\centering
\small
\caption{Robustness and fairness controls. Compute-matched direct training narrows the gap but should be interpreted separately from deployable phase-aware scheduling; reactive selectors underperform the stable schedule in this sparse small-candidate regime; scale matching helps but does not fully explain the phase-aware gain.}
\label{tab:table4}
\begin{tabularx}{\textwidth}{@{}>{\raggedright\arraybackslash}p{0.56\textwidth}>{\centering\arraybackslash}p{0.07\textwidth}>{\centering\arraybackslash}p{0.09\textwidth}>{\centering\arraybackslash}p{0.09\textwidth}>{\centering\arraybackslash}p{0.13\textwidth}@{}}
\toprule
Method & $n$ & Max $\uparrow$ & Final $\uparrow$ & Tail AUC $\uparrow$ \\
\midrule
\multicolumn{5}{@{}l}{\textbf{A. Compute-matched fairness}} \\
\texttt{direct\_oracle\_or\_direct\_best} & 8 & 0.215 & 0.022 & 0.045 \\
\texttt{direct\_oracle\_plus\_verification\_steps} & 8 & 0.339 & 0.019 & 0.023 \\
\texttt{fixed\_wu50\_hard} & 8 & 0.389 & 0.163 & 0.130 \\
\texttt{rhyve\_heldout\_selected} & 5 & 0.394 & -- & -- \\
\midrule
\multicolumn{5}{@{}l}{\textbf{B. Selector baseline}} \\
\texttt{fixed\_wu50\_hard} & 8 & 0.389 & -- & -- \\
\texttt{conservative\_periodic\_selector} & 8 & 0.051 & -- & -- \\
\texttt{moving\_average\_selector} & 8 & 0.113 & -- & -- \\
\texttt{naive\_last\_selector} & 8 & 0.092 & -- & -- \\
\texttt{direct\_oracle\_or\_direct\_best} & 8 & 0.215 & -- & -- \\
\midrule
\multicolumn{5}{@{}l}{\textbf{C. Scale control}} \\
\texttt{direct\_oracle} & 8 & 0.215 & 0.022 & 0.034 \\
\texttt{fixed\_wu50\_hard} & 8 & 0.389 & 0.163 & 0.157 \\
\texttt{direct\_oracle\_scale\_matched\_to\_early} & 8 & 0.235 & -- & -- \\
\bottomrule
\end{tabularx}

\end{table}

\subsection{Boundary Conditions and Scope}

\textsc{BallBalance} provides a dense-oracle boundary condition: all locked methods saturate the peak metric, so \texttt{consec\_max} is not discriminative. We therefore treat it as evidence that warm-up is not universally necessary rather than as another performance-dominance benchmark.

We also attempted an additional \textsc{FrankaCubeStack} sparse-manipulation pilot under the reduced optional-scope budget. All compared methods failed to reach nontrivial success, so we report it only as an all-failure boundary in the appendix rather than as a method-ranking comparison. This boundary result does not support a generalization claim, but it clarifies the scope of the current method: \textsc{RHyVE} requires a learnable task/reward setup under the available budget.

\subsection{Additional Evidence and Scope}

The appendix provides full fork-verification grids, per-seed results, tail audits, switch-operator analyses, LLM candidate-source details, selector failure taxonomy, compute-matched metrics, candidate-pool stress tests, and the \textsc{FrankaCubeStack} boundary pilot. The overall evidence supports a task-aware conclusion: \textsc{RHyVE} is strongest as a small-candidate verification-and-deployment protocol for sparse, phase-sensitive regimes. It is not a universal adaptive scheduler, not a large-pool reward-search method, and not a guarantee that warm-up improves every task.
\section{Discussion and Limitations}
\label{sec:discussion}

The main lesson is that automated reward design does not end at reward generation. Even plausible LLM-generated rewards are not immediately reliable objectives: they must be verified against the competence of the current learner and deployed with attention to when objective changes are actually justified. RHYVE is therefore best understood as a verification-informed deployment protocol rather than as a new reward generator.

The empirical scope is task-aware rather than universal. \textsc{FrankaCabinet} benefits from phase-aware deployment because the useful reward changes with competence, while \textsc{BallBalance} acts as a dense-oracle boundary condition in which warm-up is unnecessary or metric-dependent. The additional \textsc{FrankaCubeStack} pilot is an all-failure boundary under the optional budget, showing that RHyVE cannot produce meaningful rankings when the underlying task/reward setup is not learnable. These boundary cases are part of the method's diagnostic role rather than failures of a universal scheduler.

Reward switching also remains a trade-off rather than a solved subproblem. Hard switching is the strongest practical default in the locked setting, while PBRS and critic reset can help in some regimes but expose different stability--retention costs. The method therefore separates the question of whether a reward change is justified from the question of how that change should be executed.

Compute and scale must also be interpreted carefully. Compute-matched direct training narrows the gap to phase-aware deployment, confirming that verification overhead is a real fairness consideration. Reward-scale controls likewise show that scale mismatch contributes to cold-start failures. However, neither extra compute nor scale correction fully subsumes the phase-composition interpretation in the reported sparse regime.

Finally, RHYVE is a small-candidate local-verification protocol. Its conclusions depend on fork horizon, checkpoint selection, competence proxies, and candidate-family quality. Larger candidate pools will likely require uncertainty-aware ranking, coarse-to-fine verification, clustering, or more efficient candidate pruning. In practice, RHYVE should be used diagnostically: first test whether reward rankings are reliable under the current competence regime, then decide whether a single reward, a two-stage schedule, or conservative fallback is appropriate.

\section{Conclusion}
\label{sec:conclusion}

We introduced RHYVE, a competence-aware protocol for verifying and deploying generated reward hypotheses. The central finding is that reward verification can be competence-limited: a reward that appears weak from a low-competence policy can become useful later, while dense early shaping can cease to be the right objective once the learner reaches task-relevant states. On the locked \textsc{FrankaCabinet} protocol, phase-aware deployment improves peak and retained performance. Updated LLM-generated candidate experiments show that this issue also arises beyond the hand-designed reward family, but that the best deployment rule depends on the candidate pool. Controls with held-out schedule selection, conservative selectors, compute-matched direct training, scale matching, dense boundary tasks, and an all-failure extra-task pilot delimit the claim. RHYVE is not a universal scheduler or a safe-switching guarantee; it is a verification-informed deployment protocol for small candidate sets in regimes where reward utility may change with policy competence. Future reward-design systems should therefore reason not only about how to generate rewards, but also about when they become reliable and useful to deploy.

\bibliographystyle{unsrtnat}
\bibliography{references}  

\appendix
\section{Protocol, Metrics, and Reproducibility}
\label{app:protocol}

This appendix describes the reporting protocol, metric definitions, implementation details, and reproducibility conventions used throughout the paper. The goal is to make clear which results support the main claims, which results are mechanism or boundary evidence, and how the reported metrics are computed.

\subsection{Evidence Status and Reporting Protocol}
\label{app:evidence-status}

The experiments in this paper are organized by evidence status. This distinction is important because the paper combines locked main comparisons, reduced 6$\times$3090 updates, mechanism-oriented ablations, stress tests, and boundary pilots. Only the locked structured \textsc{FrankaCabinet} comparison is used for the primary sparse-task quantitative claim. Reduced-main LLM rows support candidate-family-dependent deployment behavior. Selector, compute-matched, scale, stress-test, and boundary rows are used to explain mechanisms, establish scope, or delimit the method.

\begin{table}[t]
\centering
\small
\caption{Evidence-status taxonomy used throughout the paper. Only the locked structured \textsc{FrankaCabinet} comparison supports the main sparse-task quantitative claim. Other categories explain mechanisms, establish scope, or delimit the method.}
\label{tab:tableA1}
\begin{tabularx}{0.98\textwidth}{@{}>{\raggedright\arraybackslash}p{0.24\textwidth}>{\raggedright\arraybackslash}X>{\raggedright\arraybackslash}p{0.26\textwidth}@{}}
\toprule
Status & Purpose & Example role \\
\midrule
\texttt{locked\_main\_existing} & Primary locked structured comparison used for the main sparse-task quantitative claim. & Structured \textsc{FrankaCabinet} main result \\
\texttt{reduced\_main\_6x3090} & Reduced-main LLM evidence supporting candidate-family-dependent deployment behavior. & LLM \textsc{FrankaCabinet} reduced-main rows \\
\shortstack[l]{\texttt{heldout\_test\_}\\\texttt{6x3090}} & Dev-selected schedule evaluated on held-out report seeds. & Held-out schedule audit \\
\shortstack[l]{\texttt{selector\_support\_}\\\texttt{6x3090}} & Conservative selector support evidence rather than full online-selector reproduction. & Periodic or moving-average selectors \\
\shortstack[l]{\texttt{compute\_matched\_}\\\texttt{6x3090}} & Fairness-control rows with verification-equivalent extra training budget. & Compute-matched direct training \\
\shortstack[l]{\texttt{appendix\_support\_}\\\texttt{topup}} & Mechanism-oriented top-up rows that clarify confounds without becoming main baselines. & Scale top-up controls \\
\shortstack[l]{\texttt{optional\_extra\_task\_}\\\texttt{pilot\_6x3090}} & Optional extra-task pilot used only as scope or boundary evidence. & \textsc{FrankaCubeStack} all-failure pilot \\
\shortstack[l]{\texttt{oracle\_upper\_}\\\texttt{bound}} & Non-deployable reference separated from deployable methods. & Best adaptive oracle row \\
\texttt{stress\_tests} & Scope-setting experiments outside the strongest claim regime. & Candidate-pool scaling \\
\shortstack[l]{\texttt{historical\_}\\\texttt{legacy}} & Older breadth or pilot evidence retained only for traceability. & Legacy breadth rows \\
\shortstack[l]{\texttt{failed\_or\_}\\\texttt{blocked}} & Incomplete, flat, or blocked evidence not used for ranking claims. & Flat or blocked appendix entries \\
\bottomrule
\end{tabularx}

\end{table}

For each experimental block, we retain a manifest that records the task, reward-hypothesis family, seed list, training budget, fork horizon, checkpoint set, switch rule, switch operator, metrics, and evidence status. The main paper reports only compact summaries. Full per-seed values, paired tests, and diagnostic tables are included in the appendix.

\paragraph{Locked comparisons.}
A locked comparison is defined before aggregation by fixing the set of methods, seeds, reward family, evaluation protocol, and reporting metrics. For \textsc{FrankaCabinet}, the locked main comparison uses eight seeds and compares direct oracle deployment, one-shot oracle deployment, fixed warm-up schedules, and value-aligned PBRS. For \textsc{BallBalance}, the locked boundary comparison uses the same dense-oracle reward family and reports peak saturation together with AUC and tail metrics.

\paragraph{Support and stress experiments.}
Appendix-support experiments are not treated as additional main baselines. For example, reward-scale controls mix locked and mechanism-only evidence, and are therefore explicitly marked by evidence status. Candidate-pool scaling is reported as a stress test because it probes whether the small-candidate local-verification assumption continues to hold as $K$ increases.

\subsection{Metric Definitions}
\label{app:metrics}

The main paper reports both peak and tail metrics. This is necessary because a method can achieve a high transient success value and later collapse. We therefore recompute terminal and tail metrics from full learning curves whenever possible.

\begin{table}[t]
\centering
\small
\caption{Metric definitions. Peak metrics summarize whether a method ever reaches high performance, while tail metrics measure whether performance is retained near the end of training.}
\label{tab:tableA2}
\resizebox{0.96\textwidth}{!}{\begin{tabular}{p{0.20\linewidth}p{0.50\linewidth}p{0.22\linewidth}}
\toprule
Metric & Definition & Primary use \\
\midrule
\texttt{consec\_max} & Maximum consecutive-success score observed during training. & Peak performance \\
\texttt{consec\_auc} & Area under the consecutive-success curve over the full training trajectory. & Training efficiency \\
Recomputed final success & Final success value recomputed from the full logged learning curve. & Terminal performance \\
Last-5 mean & Mean success over the final five evaluation points. & Tail stability \\
Tail AUC & Area under the success curve over the final training window. & Retained late-stage performance \\
Best-checkpoint success & Maximum evaluation success over all saved checkpoints. & Best attainable checkpoint performance \\
Collapse indicator & Indicator for high best-checkpoint success with weak terminal retention. & Detecting transient spikes \\
\texttt{critic\_peak\_abs} & Peak absolute critic magnitude or critic-instability proxy. & Switching stability \\
$|\Delta \bar r|$ & Absolute mean reward shift induced by a reward switch. & Reward shock / scale mismatch \\
\bottomrule
\end{tabular}
}
\end{table}

\paragraph{Recomputed terminal metrics.}
Some result files contain summary fields produced by different logging paths. To avoid inconsistent terminal reporting, we recompute final and tail metrics directly from the full learning curves. The main text therefore uses recomputed final success, last-5 mean, and tail AUC when discussing terminal retention. Older summary fields are retained only for traceability and are not used as main evidence.

\paragraph{Collapse rate.}
We define collapse as a diagnostic rather than a primary optimization objective. A run is marked as collapsed when it reaches a nontrivial best checkpoint but has low terminal retention. In our analysis scripts, the default rule is
\[
\mathbb{I}_{\mathrm{collapse}}
=
\mathbf{1}
\left[
\mathrm{best\_success}\geq \tau_{\mathrm{best}}
\;\wedge\;
\mathrm{last5}\leq \tau_{\mathrm{tail}}
\right],
\]
where $\tau_{\mathrm{best}}$ and $\tau_{\mathrm{tail}}$ are task-dependent thresholds. The thresholds are reported in the corresponding audit tables.

\paragraph{Bootstrap intervals and paired tests.}
For aggregate tables, we report means, standard deviations, and bootstrap confidence intervals over seeds. For locked paired comparisons, we use exact sign-flip permutation tests when the number of paired seeds is small. We additionally report paired standardized effect sizes. Because the main sparse-control task exhibits high seed variance, we interpret statistical tests together with effect magnitude, tail metrics, and mechanism-consistent ablations.

\subsection{Tasks, Reward Families, and Seed Protocol}
\label{app:tasks-seeds}

Table~\ref{tab:tableA3} summarizes the tasks and their role in the paper. \textsc{FrankaCabinet} is the main sparse, phase-sensitive manipulation task. \textsc{BallBalance} is a dense-oracle boundary condition. \textsc{CartPole} and \textsc{Ant} are appendix-only breadth or stress tasks. \textsc{FrankaCubeStack} is an optional extra-task pilot; because all compared methods failed to reach nontrivial success under the reduced optional budget, it is reported as an all-failure boundary rather than a method-ranking benchmark.

\begin{table}[t]
\centering
\small
\caption{Task roles and reporting status. The main claims are based on \textsc{FrankaCabinet} locked results and \textsc{BallBalance} boundary evidence. Other tasks are used for breadth, scope, or all-failure boundary evidence.}
\label{tab:tableA3}
\resizebox{0.96\textwidth}{!}{\begin{tabular}{p{0.18\linewidth}p{0.22\linewidth}p{0.34\linewidth}p{0.18\linewidth}}
\toprule
Task & Regime & Role in paper & Evidence status \\
\midrule
\textsc{FrankaCabinet} & Sparse, phase-sensitive manipulation & Main competence-aware verification and phase-aware deployment task & \texttt{locked\_main\_existing} \\
\textsc{BallBalance} & Dense-oracle boundary & Boundary evidence where warm-up is not meaningfully ranked by peak success & \texttt{historical\_legacy} / boundary \\
\textsc{CartPole} & Appendix breadth / stress task & Breadth and candidate-pool scaling boundary evidence & \texttt{historical\_legacy} / \texttt{stress\_tests} \\
\textsc{Ant} & Appendix breadth / negative-scope task & Weak or fragile breadth evidence & \texttt{historical\_legacy} \\
\textsc{FrankaCubeStack} & Optional extra-task pilot & All-failure boundary; not used for method ranking & \texttt{optional\_extra\_task\_pilot\_6x3090} \\
\bottomrule
\end{tabular}
}
\end{table}

\paragraph{Reward-hypothesis families.}
The main experiments use small candidate sets, typically $K=3$. For \textsc{FrankaCabinet}, the reward family contains an early dense bootstrap reward, a later task-faithful oracle-like reward, and a later alternative. For \textsc{BallBalance}, the reward family contains an early speed-biased reward, an oracle-like position-speed reward, and a distance-based alternative. LLM-generated reward families are evaluated separately in the generalization and scope section of the appendix.

\paragraph{Seed protocol.}
The locked \textsc{FrankaCabinet} comparison uses the seed set
\[
\{42, 123, 456, 789, 999, 1234, 2025, 3031\}.
\]
The locked \textsc{BallBalance} boundary comparison uses a fixed subset of seeds from the same experimental suite. Some appendix-support experiments use smaller seed sets because they serve as diagnostic or stress evidence rather than main claims. Each table reports its own $n$.

\subsection{Training and Verification Configuration}
\label{app:training-config}

The main experiments use actor-critic policy optimization with PPO-style updates~\cite{schulman2017proximal}. The simulator and task implementation follow Isaac Gym-style GPU-accelerated control environments~\cite{makoviychuk2021isaac}. Table~\ref{tab:app-training-config} gives the high-level configuration summary. Exact hyperparameters and reward definitions are provided in the supplementary configuration files.

\begin{table}[t]
\centering
\small
\setlength{\tabcolsep}{4pt}
\caption{High-level configuration summary. Exact low-level hyperparameters are stored in the released configuration files.}
\label{tab:app-training-config}

\begin{tabular}{p{0.20\linewidth}p{0.20\linewidth}p{0.24\linewidth}p{0.24\linewidth}}
\toprule
Experiment block & Task & Candidate family & Main reporting role \\
\midrule
Locked main comparison
& \textsc{FrankaCabinet}
& Small structured triplet.
& Main phase-aware deployment result. \\
\midrule
Boundary comparison
& \textsc{BallBalance}
& Small structured triplet.
& Dense-oracle boundary condition. \\
\midrule
Fork verification sweep
& \textsc{FrankaCabinet}, \textsc{BallBalance}
& Same task-specific triplets.
& Competence and fork-horizon reliability. \\
\midrule
Reward-scale controls
& \textsc{FrankaCabinet}
& Early/late reward pair and scale variants.
& Mechanism and confound analysis. \\
\midrule
Switch-operator grid
& \textsc{FrankaCabinet}
& Early/late reward pair.
& Stability--retention trade-off. \\
\midrule
Source generalization
& \textsc{FrankaCabinet}, \textsc{BallBalance}, \textsc{Cartpole}
& Hand-crafted and LLM-generated families.
& Phenomenon generalization. \\
\midrule
Pool scaling
& \textsc{FrankaCabinet}, \textsc{Cartpole}
& Expanded candidate pools.
& Stress test and scope. \\
\bottomrule
\end{tabular}
\end{table}

\paragraph{Fork verification configuration.}
At each probed checkpoint $z_t$, every candidate reward hypothesis is evaluated by cloning the same policy, critic, and optimizer state, training each clone for a short fork horizon, and scoring the resulting policies with a common downstream metric. The fork horizon is varied in the reliability analysis to test whether winner identity is robust to the verification budget.

\paragraph{Checkpoint spacing.}
The checkpoint set is task-dependent. Sparse tasks use checkpoints that cover early, middle, and late training phases. Boundary tasks such as \textsc{BallBalance} use earlier checkpoints because they reach high success much faster. The exact checkpoint list for each experiment is recorded in the manifest.

\paragraph{Computational complexity.}
If $M$ checkpoints are probed, $K$ reward hypotheses are compared, and each fork uses horizon $L$, the verification overhead scales as
\[
O(MKL).
\]
The intended regime of \textsc{RHyVE} is small $K$, sparse $M$, and short-to-medium $L$. Candidate-pool scaling experiments reported later in the appendix show that larger $K$ increases both verification cost and ranking difficulty.

\subsection{Algorithms}
\label{app:algorithms}

For completeness, we give the procedural form of the two main operations used by \textsc{RHyVE}. We write the algorithms in text form to avoid dependency on additional algorithm packages.

\paragraph{Algorithm A1: Shared-checkpoint fork verification.}

\begin{enumerate}
    \item \textbf{Input:} candidate reward hypotheses $\mathcal{H}=\{h_1,\ldots,h_K\}$, checkpoint set $\mathcal{T}$, fork horizon $L$, evaluation function $\operatorname{Eval}$.
    \item For each checkpoint $t\in\mathcal{T}$:
    \begin{enumerate}
        \item Load the shared learner state $z_t=(\pi_t,V_t,\Omega_t)$.
        \item Compute or record the task competence proxy $c_t$.
        \item For each reward hypothesis $h_k\in\mathcal{H}$:
        \begin{enumerate}
            \item Clone the learner state $z_t$.
            \item Continue training the clone for $L$ updates using reward $r_{h_k}$.
            \item Evaluate the resulting forked policy and record $J_t^{(L)}(h_k)$.
        \end{enumerate}
        \item Compute the local winner
        \[
        \hat h_t = \arg\max_{h_k\in\mathcal{H}} J_t^{(L)}(h_k).
        \]
        \item Compute the winner margin
        \[
        m_t = J_t^{(L)}(\hat h_t) -
        \max_{h_j\neq \hat h_t}J_t^{(L)}(h_j).
        \]
        \item Store $(t,c_t,\hat h_t,m_t,\{J_t^{(L)}(h_k)\}_{k=1}^K)$.
    \end{enumerate}
    \item \textbf{Output:} phase profile $\mathcal{P}$.
\end{enumerate}

\paragraph{Algorithm A2: Phase-aware deployment.}

\begin{enumerate}
    \item \textbf{Input:} phase profile $\mathcal{P}$, reward hypotheses $\mathcal{H}$, optional switch-operator set.
    \item Identify checkpoints that satisfy the verification-informative criterion, using margin, agreement, and entropy diagnostics.
    \item If a single reward hypothesis wins consistently from the first informative checkpoint onward, deploy that reward throughout training.
    \item If an early winner $h^{(1)}$ is later overtaken by a stable later winner $h^{(2)}$, construct a two-stage schedule:
    \[
    r_t =
    \begin{cases}
    r_{h^{(1)}} & t < t_s,\\
    r_{h^{(2)}} & t \ge t_s.
    \end{cases}
    \]
    \item Choose the switch operator:
    \begin{enumerate}
        \item use hard switching as the default practical operator;
        \item use value-aligned shaping only as a conditional mechanism;
        \item use critic reset only as a diagnostic or trade-off operator.
    \end{enumerate}
    \item If no stable phase structure is observed, abstain from aggressive switching and retain a conservative deployment rule.
    \item \textbf{Output:} deployed training schedule and switch-operator choice.
\end{enumerate}

\subsection{Artifact and Table-Generation Discipline}
\label{app:artifacts}

All paper-ready tables and figures are generated from CSV summaries that are themselves derived from per-seed raw logs. To avoid accidental mixing of reporting modes, each generated table records its source file and evidence status. Before submission, the following consistency checks are applied:

\begin{enumerate}
    \item all main-text tables use only \texttt{locked\_main} or explicitly marked boundary evidence;
    \item all appendix-support and stress-test results are labeled as such;
    \item no table mixes locked and appendix evidence without an explicit evidence-status column;
    \item all terminal metrics in the main text use recomputed final or tail metrics rather than inconsistent summary fields;
    \item all incomplete values are removed before final submission.
\end{enumerate}

Table~\ref{tab:app-artifact-map} summarizes the main artifact groups used to generate the paper-ready results.

\begin{table}[tbp]
\centering
\small
\setlength{\tabcolsep}{3.2pt}
\caption{Artifact groups used for paper-ready tables and figures. Paths are represented by logical artifact names rather than machine-specific absolute paths.}
\label{tab:app-artifact-map}

\begin{tabular}{p{0.30\linewidth}p{0.56\linewidth}}
\toprule
Artifact group & Purpose \\
\midrule
Experiment manifests
& Define tasks, methods, seeds, reward families, switch rules, and evidence status. \\
\midrule
Per-seed raw logs
& Store learning curves, checkpoint metrics, critic diagnostics, and reward-shift statistics. \\
\midrule
Locked result CSVs
& Generate main comparison tables and locked paired tests. \\
\midrule
Appendix-support CSVs
& Generate reward-scale, switching, verification, trigger, and source-generalization tables. \\
\midrule
Stress-test CSVs
& Generate candidate-pool scaling and compute--reliability plots. \\
\midrule
Paper-ready table scripts
& Convert source CSVs into LaTeX tables with evidence-status annotations. \\
\midrule
Paper-ready figure scripts
& Generate learning curves, heatmaps, trade-off plots, and reliability figures. \\
\bottomrule
\end{tabular}
\end{table}
\section{Full Verification and Main-Task Results}
\label{app:full-results}

This appendix provides the full evidence behind the main experimental claims. The following subsections expand the competence-limited verification analysis, report the locked \textsc{FrankaCabinet} results with per-method and statistical details, and document the dense-oracle \textsc{BallBalance} boundary condition.

\subsection{Full Competence and Fork-Horizon Reliability}
\label{app:verification-full}

The main text summarizes the central verification phenomenon: reward rankings obtained from fork verification are not uniformly reliable throughout training. Here we provide the full checkpoint and fork-horizon results. For each task, we evaluate candidate reward hypotheses from shared checkpoints using multiple fork horizons. We record the modal winner, repeated-fork top-1 reliability, winner margin, margin lower confidence bound, and whether the checkpoint satisfies the verification-informative criterion.

\paragraph{Verification-informative criterion.}
Unless otherwise stated, a checkpoint is treated as verification-informative when the modal winner is sufficiently stable across repeated forks and has a positive margin lower bound:
\[
\operatorname{Rel}_t \geq 0.75
\qquad\text{and}\qquad
\operatorname{LCB}_{95}(m_t) > 0.01 .
\]
For \textsc{FrankaCabinet}, margins are small because success metrics are normalized, and the same criterion is used only as a diagnostic threshold rather than a universal rule. The important qualitative pattern is that early checkpoints are less reliable and later checkpoints become more informative under sufficiently long fork horizons.

\begin{table}[tbp]
\centering
\small
\setlength{\tabcolsep}{4pt}
\caption{First reliable checkpoint by task and fork horizon. Reliability depends on both policy competence and verification horizon.}
\label{tab:app-first-reliable}
\begin{tabular}{lcc}
\toprule
Task & Fork horizon & First reliable checkpoint \\
\midrule
\textsc{BallBalance} & 50  & 30 \\
\textsc{BallBalance} & 100 & 300 \\
\textsc{BallBalance} & 200 & 300 \\
\textsc{BallBalance} & 400 & 30 \\
\midrule
\textsc{FrankaCabinet} & 50  & 1500 \\
\textsc{FrankaCabinet} & 100 & 1500 \\
\textsc{FrankaCabinet} & 200 & 1000 \\
\textsc{FrankaCabinet} & 400 & 1000 \\
\bottomrule
\end{tabular}
\end{table}

\begin{figure}[t]
    \centering
    \includegraphics[width=0.98\textwidth]{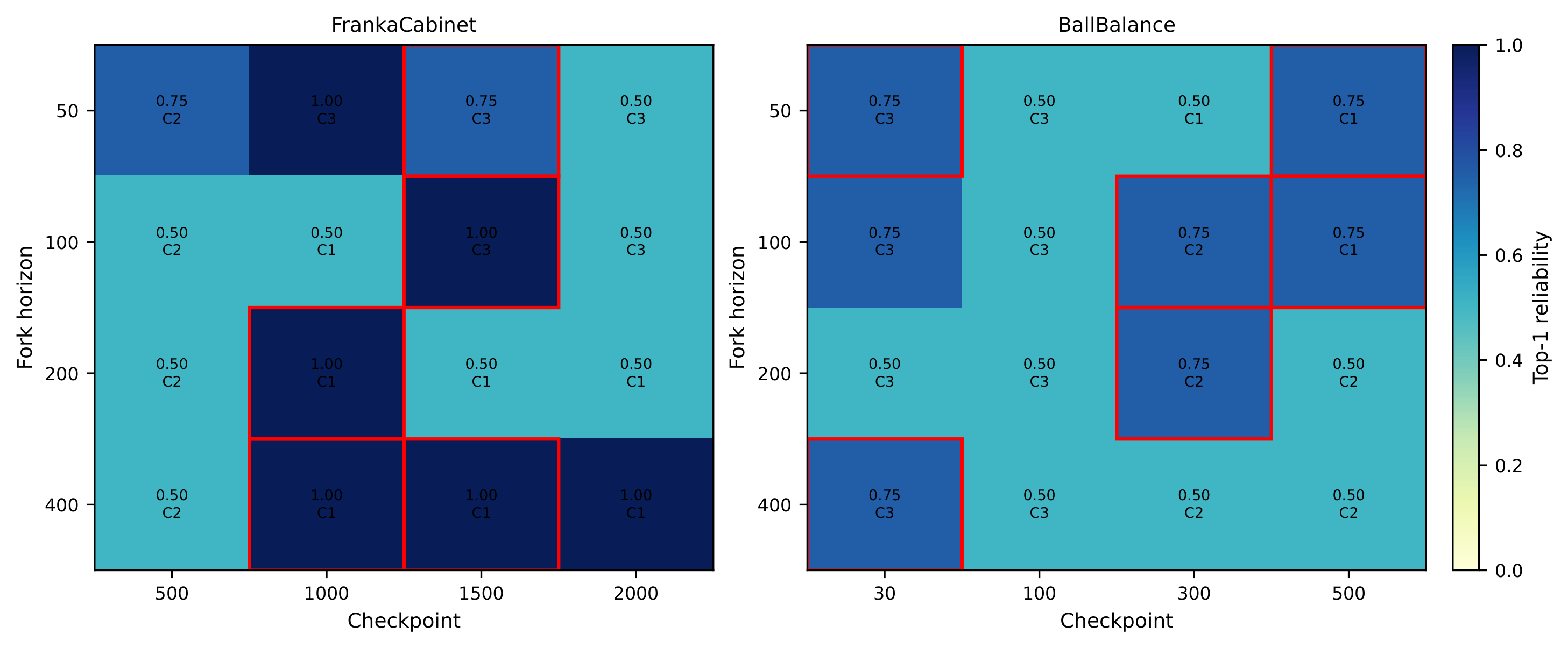}
    \caption{Full fork-verification reliability. Each cell aggregates repeated forks from a shared checkpoint and reports modal winner identity, top-1 reliability, and winner margin. The reliable region appears only after sufficient policy competence and depends on fork horizon.}
    \label{fig:app-verification-heatmap}
\end{figure}

\begin{table}[t]
\centering
\small
\setlength{\tabcolsep}{3.2pt}
\caption{Full \textsc{FrankaCabinet} fork-verification grid. The table reports modal winner and top-1 reliability for each checkpoint and fork horizon. Margins and entropy are reported in the released CSV.}
\label{tab:app-franka-verification-grid}
\begin{tabular}{cccccc}
\toprule
Checkpoint & Fork horizon & Modal winner & Top-1 reliability & Margin mean & Informative \\
\midrule
500  & 50  & \texttt{cand\_b7d2e641} & 0.75 & 0.00017 & no \\
500  & 100 & \texttt{cand\_b7d2e641} & 0.50 & 0.00007 & no \\
500  & 200 & \texttt{cand\_b7d2e641} & 0.50 & 0.00057 & no \\
500  & 400 & \texttt{cand\_b7d2e641} & 0.50 & 0.01552 & no \\
\midrule
1000 & 50  & \texttt{cand\_bc49b15b} & 1.00 & 0.01419 & no \\
1000 & 100 & \texttt{cand\_6ef57213} & 0.50 & 0.00755 & no \\
1000 & 200 & \texttt{cand\_6ef57213} & 1.00 & 0.09161 & yes \\
1000 & 400 & \texttt{cand\_6ef57213} & 1.00 & 0.08832 & yes \\
\midrule
1500 & 50  & \texttt{cand\_bc49b15b} & 0.75 & 0.05101 & yes \\
1500 & 100 & \texttt{cand\_bc49b15b} & 1.00 & 0.05602 & yes \\
1500 & 200 & \texttt{cand\_6ef57213} & 0.50 & 0.02488 & no \\
1500 & 400 & \texttt{cand\_6ef57213} & 1.00 & 0.09381 & yes \\
\midrule
2000 & 50  & \texttt{cand\_bc49b15b} & 0.50 & 0.04170 & no \\
2000 & 100 & \texttt{cand\_bc49b15b} & 0.50 & 0.01654 & no \\
2000 & 200 & \texttt{cand\_6ef57213} & 0.50 & 0.00329 & no \\
2000 & 400 & \texttt{cand\_6ef57213} & 1.00 & 0.04448 & no \\
\bottomrule
\end{tabular}
\end{table}

The \textsc{FrankaCabinet} grid supports two observations. First, the earliest checkpoint is not verification-informative despite occasionally producing a modal winner. Second, later checkpoints become informative only under sufficiently reliable fork horizons. This justifies the main text's competence-aware interpretation: reward ranking is not a static property of the candidate set, but depends on the current learner and the verification budget.

\begin{table}[t]
\centering
\small
\setlength{\tabcolsep}{3.2pt}
\caption{Full \textsc{BallBalance} fork-verification grid. \textsc{BallBalance} reaches high task competence quickly, and peak metrics saturate, making the task a boundary condition rather than a main performance benchmark.}
\label{tab:app-ballbalance-verification-grid}
\begin{tabular}{cccccc}
\toprule
Checkpoint & Fork horizon & Modal winner & Top-1 reliability & Margin mean & Informative \\
\midrule
30  & 50  & \texttt{cand\_bb\_speed\_bias} & 0.75 & 54.10 & yes \\
30  & 100 & \texttt{cand\_bb\_speed\_bias} & 0.75 & 52.59 & no \\
30  & 200 & \texttt{cand\_bb\_speed\_bias} & 0.50 & 39.96 & no \\
30  & 400 & \texttt{cand\_bb\_speed\_bias} & 0.75 & 26.16 & yes \\
\midrule
100 & 50  & \texttt{cand\_bb\_speed\_bias} & 0.50 & 30.01 & no \\
100 & 100 & \texttt{cand\_bb\_speed\_bias} & 0.50 & 19.16 & no \\
100 & 200 & \texttt{cand\_bb\_speed\_bias} & 0.50 & 26.27 & no \\
100 & 400 & \texttt{cand\_bb\_speed\_bias} & 0.50 & 25.82 & no \\
\midrule
300 & 50  & \texttt{cand\_bb\_dist\_only} & 0.50 & 17.12 & no \\
300 & 100 & \texttt{cand\_bb\_pos\_speed} & 0.75 & 37.02 & yes \\
300 & 200 & \texttt{cand\_bb\_pos\_speed} & 0.75 & 46.08 & yes \\
300 & 400 & \texttt{cand\_bb\_pos\_speed} & 0.50 & 30.73 & no \\
\midrule
500 & 50  & \texttt{cand\_bb\_dist\_only} & 0.75 & 22.98 & yes \\
500 & 100 & \texttt{cand\_bb\_dist\_only} & 0.75 & 18.56 & yes \\
500 & 200 & \texttt{cand\_bb\_pos\_speed} & 0.50 & 8.45  & no \\
500 & 400 & \texttt{cand\_bb\_pos\_speed} & 0.50 & 17.03 & no \\
\bottomrule
\end{tabular}
\end{table}

The \textsc{BallBalance} grid shows that reward winner identity can still change across checkpoints, but the downstream peak metric is less discriminative because all main deployment methods reach the ceiling. We therefore use \textsc{BallBalance} primarily to show that warm-up is not universally needed when the oracle-like reward is already dense.

\subsection{FrankaCabinet Locked Main Results}
\label{app:franka-full}

This section expands the locked \textsc{FrankaCabinet} comparison from the main text. The locked comparison uses eight seeds:
\[
\{42, 123, 456, 789, 999, 1234, 2025, 3031\}.
\]
All methods use the same task, reward-family protocol, training budget, and evaluation pipeline. Table~\ref{tab:app-franka-locked-full} reports the aggregate peak results.

\begin{table}[t]
\centering
\small
\setlength{\tabcolsep}{4pt}
\caption{Locked \textsc{FrankaCabinet} peak results. \texttt{wu=50} hard switching achieves the highest mean \texttt{consec\_max}.}
\label{tab:app-franka-locked-full}
\begin{tabular}{lcccc}
\toprule
Method & $n$ & \texttt{consec\_max} mean & Std. & 95\% CI \\
\midrule
Direct oracle & 8 & 0.21498 & 0.28109 & [0.05100, 0.41471] \\
One-shot oracle & 8 & 0.22815 & 0.27984 & [0.06691, 0.42390] \\
Fixed \texttt{wu=50} hard & 8 & \textbf{0.38894} & 0.22560 & [0.23248, 0.53150] \\
Fixed \texttt{wu=100} hard & 8 & 0.23263 & 0.19816 & [0.11569, 0.37862] \\
\texttt{wu=50} + PBRS & 8 & 0.31341 & 0.27950 & [0.13502, 0.49579] \\
\bottomrule
\end{tabular}
\end{table}

\begin{table}[t]
\centering
\small
\setlength{\tabcolsep}{3.5pt}
\caption{Per-seed \textsc{FrankaCabinet} \texttt{consec\_max} values for locked main methods. These values are included to make the seed variance visible.}
\label{tab:app-franka-perseed}
\begin{tabular}{lcccccccc}
\toprule
Method & 42 & 123 & 456 & 789 & 999 & 1234 & 2025 & 3031 \\
\midrule
Direct oracle
& 0.251 & 0.026 & 0.146 & 0.000 & 0.003 & 0.602 & 0.689 & 0.002 \\
One-shot oracle
& 0.251 & 0.026 & 0.251 & 0.000 & 0.003 & 0.602 & 0.689 & 0.002 \\
Fixed \texttt{wu=50} hard
& 0.033 & 0.505 & 0.602 & 0.180 & 0.655 & 0.380 & 0.210 & 0.546 \\
Fixed \texttt{wu=100} hard
& 0.634 & 0.206 & 0.205 & 0.271 & 0.233 & 0.306 & 0.004 & 0.002 \\
\texttt{wu=50} + PBRS
& 0.663 & 0.013 & 0.451 & 0.341 & 0.389 & 0.650 & 0.001 & 0.000 \\
\bottomrule
\end{tabular}
\end{table}

\paragraph{Tail-metric audit.}
The main text reports recomputed final and tail metrics rather than older summary fields. Table~\ref{tab:app-franka-tail} summarizes the tail audit. These values are computed from full learning curves and are used to distinguish transient peaks from retained performance.

\begin{table}[t]
\centering
\small
\setlength{\tabcolsep}{4pt}
\caption{Tail-metric audit for locked \textsc{FrankaCabinet} runs. Final values are recomputed from the full learning curves and are used to distinguish transient peaks from retained performance.}
\label{tab:app-franka-tail}
\begin{tabular}{lcc}
\toprule
Method & $n$ & Recomputed final $\uparrow$ \\
\midrule
Direct oracle & 8 & 0.0223 \\
One-shot oracle & 8 & 0.0216 \\
Fixed \texttt{wu=50} hard & 8 & \textbf{0.1634} \\
Fixed \texttt{wu=100} hard & 8 & 0.0893 \\
\texttt{wu=50} + PBRS & 8 & 0.0364 \\
\bottomrule
\end{tabular}
\end{table}

Additional last-5, tail-AUC, and collapse diagnostics are visualized in Figure~\ref{fig:app-franka-tail} when available from the audit logs.

The tail audit strengthens the main result. The \texttt{wu=50} hard-switch schedule is not only the strongest peak method; it also improves recomputed terminal and tail metrics over direct and one-shot deployment. Therefore the main result is not merely a transient success spike.

\paragraph{Paired tests.}
Table~\ref{tab:app-franka-paired} reports paired tests for the main peak metric. Additional paired tests for final, last-5, tail AUC, and collapse diagnostics are generated from the paper-ready audit files.

\begin{table}[tbp]
\centering
\small
\setlength{\tabcolsep}{4pt}
\caption{Paired sign-flip tests for \textsc{FrankaCabinet} \texttt{consec\_max}. Effect sizes are moderate, but high seed variance prevents strong significance claims on the peak metric.}
\label{tab:app-franka-paired}
\begin{tabular}{lccc}
\toprule
Comparison & Mean diff. & $p$ & $d_z$ \\
\midrule
\texttt{wu50} vs direct & 0.174 & 0.281 & 0.408 \\
\texttt{wu50} vs one-shot & 0.161 & 0.305 & 0.384 \\
\texttt{wu50} vs \texttt{wu100} & 0.156 & 0.258 & 0.426 \\
\bottomrule
\end{tabular}
\end{table}

\begin{figure}[t]
    \centering
    \includegraphics[width=0.98\textwidth]{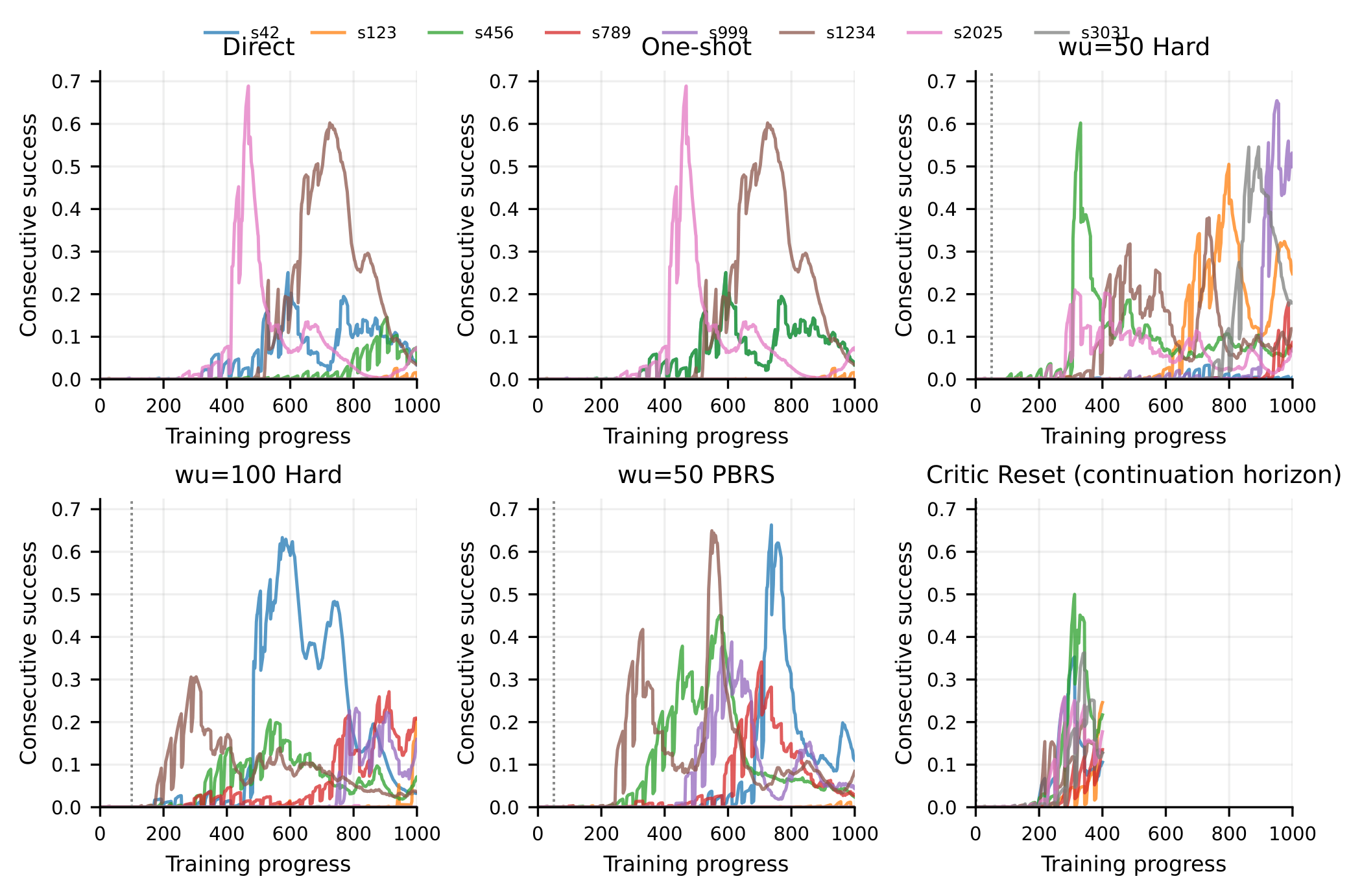}
    \caption{Per-seed learning curves for locked \textsc{FrankaCabinet} runs. The figure visualizes the high seed variance and shows why we report both peak and tail metrics.}
    \label{fig:app-franka-spaghetti}
\end{figure}

\begin{figure}[tbp]
    \centering
    \includegraphics[width=\linewidth]{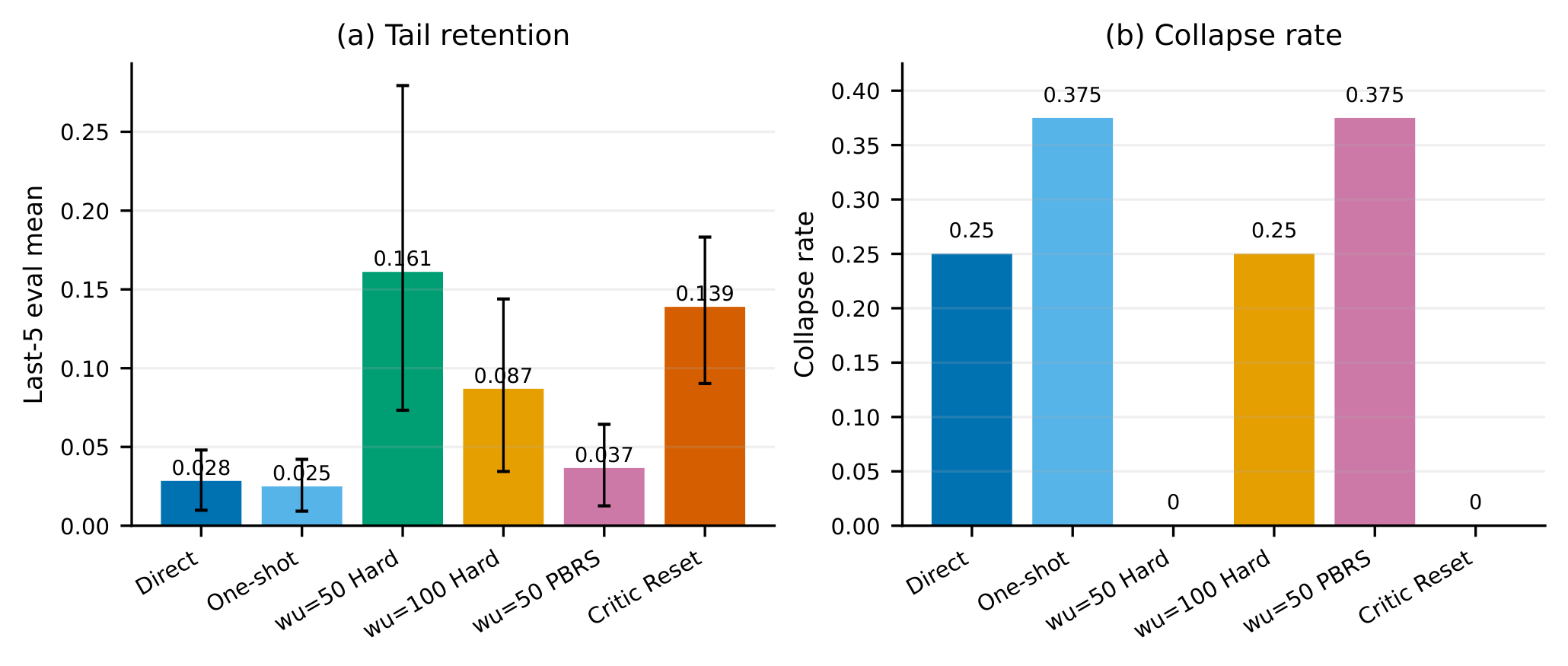}
    \caption{Tail retention diagnostics for locked \textsc{FrankaCabinet} runs. The \texttt{wu=50} schedule improves recomputed terminal and tail behavior relative to direct and one-shot deployment.}
    \label{fig:app-franka-tail}
\end{figure}

\subsection{BallBalance Boundary Details}
\label{app:ballbalance-full}

\textsc{BallBalance} is used as a dense-oracle boundary condition. In the locked comparison, all methods reach the peak ceiling, so \texttt{consec\_max} alone is not informative.

\begin{table}[tbp]
\centering
\small
\setlength{\tabcolsep}{4pt}
\caption{Locked \textsc{BallBalance} peak results. All methods saturate the peak metric, so this table should not be used for ranking methods.}
\label{tab:app-ballbalance-peak}
\begin{tabular}{lccc}
\toprule
Method & $n$ & \texttt{consec\_max} mean & Std. \\
\midrule
Direct oracle (wu=0) & 5 & 499.0 & 0.0 \\
Fixed \texttt{wu=50} hard switch & 5 & 499.0 & 0.0 \\
Fixed \texttt{wu=100} hard switch & 5 & 499.0 & 0.0 \\
\bottomrule
\end{tabular}
\end{table}

The available locked boundary evidence shows that all methods reach the peak ceiling of 499. Therefore, this task is used to establish a dense-oracle boundary condition rather than to rank methods by peak performance.
\section{Mechanism and Ablation Analyses}
\label{app:mechanism-ablation}

This appendix provides the mechanism and ablation evidence behind the main experimental interpretation. The following subsections study reward-scale controls, analyze switching operators as stability--retention trade-offs, and report selector and trigger ablations, including failure modes of reactive alternatives.

\subsection{Reward Scale and Phase-Composition Controls}
\label{app:scale-controls}

A possible alternative explanation for the main result is that direct oracle deployment fails only because of reward-scale mismatch or reward shock. To test this, we compare direct oracle deployment, scale-normalized direct deployment, scale-matched direct deployment, reward interpolation, early-only training, late-only training, and the phase-aware \texttt{wu=50} hard-switch schedule.

Table~\ref{tab:app-scale-compact} retains only the updated scale-topup summary aligned with the revised main-text control semantics. We do not keep the older appendix tables that still reported stale scale-matched direct rows from the earlier appendix snapshot.

\begin{table}[t]
\centering
\small
\setlength{\tabcolsep}{4pt}
\caption{Updated reward-scale top-up summary on \textsc{FrankaCabinet}. These rows are \texttt{appendix\_support\_topup} evidence and replace the older stale appendix scale aggregates.}
\label{tab:app-scale-compact}
\resizebox{0.98\textwidth}{!}{\begin{tabular}{lcccccl}
\toprule
Method & $n$ & Max $\uparrow$ & Std. & 95\% CI low & 95\% CI high & Status \\
\midrule
\texttt{direct\_oracle\_running\_norm} & 5 & 0.078 & 0.170 & 0.001 & 0.230 & \texttt{appendix\_support\_topup} \\
\texttt{direct\_oracle\_scale\_matched\_to\_early} & 8 & 0.235 & 0.281 & 0.074 & 0.432 & \texttt{appendix\_support\_topup} \\
\texttt{early\_dense\_only} & 5 & 0.013 & 0.025 & 0.001 & 0.036 & \texttt{appendix\_support\_topup} \\
\texttt{late\_reward\_only} & 5 & 0.013 & 0.026 & 0.001 & 0.036 & \texttt{appendix\_support\_topup} \\
\bottomrule
\end{tabular}
}
\end{table}

The updated scale-control appendix still supports the same narrow interpretation: reward scale is a real confound, but scale correction alone does not subsume the phase-aware deployment result reported in the main text.

\begin{figure}[t]
    \centering
    \includegraphics[width=\linewidth]{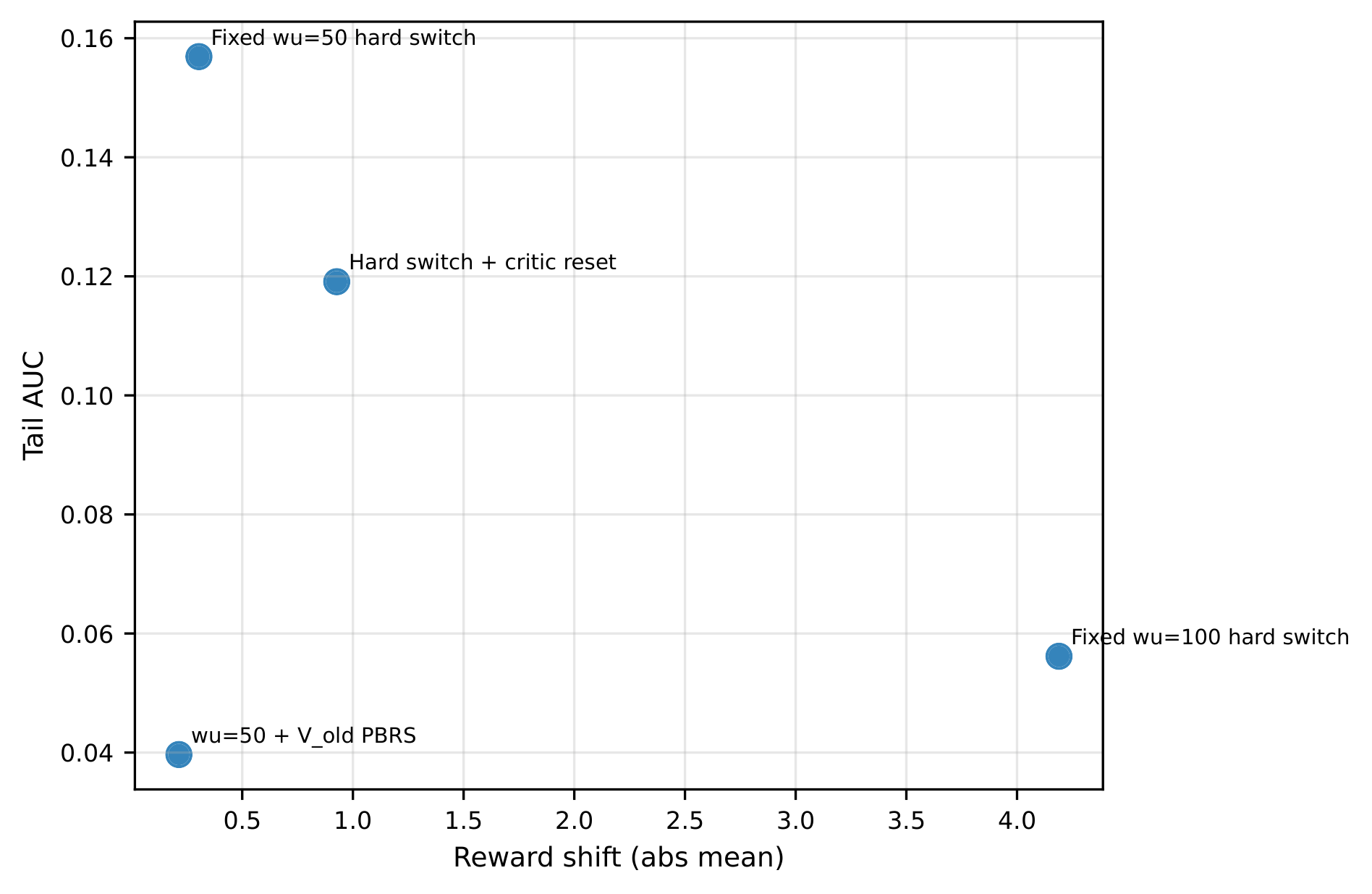}
    \caption{Reward-shift diagnostic. Reward-scale and reward-shock effects influence training stability, but they do not fully account for the phase-aware deployment gain.}
    \label{fig:app-reward-shift}
\end{figure}

\paragraph{Interpretation.}
These controls should be read as mechanism evidence rather than as additional main baselines. They rule out a simple explanation in which warm-up only acts as reward-scale normalization. Scale and shock matter, but the failure of early-only and late-only training shows that correct phase composition is also necessary.

\subsection{Switching Operator and Critic-Maturity Analysis}
\label{app:switch-controls}

We next analyze how the reward switch is executed. The main paper emphasizes that switching safety is not solved by a universal operator. Here we provide the full switching summary and the operator-by-epoch grid.

Table~\ref{tab:app-safe-locked} reports the locked \textsc{FrankaCabinet} safe-switching summary. PBRS is not a universal improvement, and critic reset improves terminal retention in some settings but induces much larger critic peaks.

\begin{table}[t]
\centering
\small
\setlength{\tabcolsep}{4pt}
\caption{Locked \textsc{FrankaCabinet} switching summary. Final values use recomputed terminal metrics. No operator uniformly dominates across metrics; the \texttt{wu=50} hard switch gives the strongest peak and recomputed final performance with low critic instability, whereas critic reset trades retention for high critic instability.}
\label{tab:app-safe-locked}
\begin{tabular}{lccccc}
\toprule
Method & $n$ & Max $\uparrow$ & Final $\uparrow$ & Critic peak $\downarrow$ & $|\Delta \bar r|$ \\
\midrule
Direct oracle & 8 & 0.215 & 0.022 & 0.480 & -- \\
Fixed \texttt{wu=50} hard & 8 & \textbf{0.389} & \textbf{0.163} & 0.064 & 0.304 \\
Fixed \texttt{wu=100} hard & 8 & 0.233 & 0.089 & \textbf{0.019} & 4.189 \\
\texttt{wu=50} + V-old PBRS & 8 & 0.313 & 0.036 & 0.077 & 0.214 \\
Hard switch + critic reset & 8 & 0.258 & 0.142 & 3.113 & 0.927 \\
\bottomrule
\end{tabular}
\end{table}

\paragraph{Operator-by-epoch grid.}
Table~\ref{tab:app-switch-grid} reports the multiseed switch-operator grid. We evaluate hard switching, PBRS with $\Phi=V_{\mathrm{old}}$, and critic reset at three switch epochs. The best operator depends on both the target metric and the switch timing. This reinforces the stability--retention interpretation: no switch operator dominates across all regimes.

\begin{table}[t]
\centering
\small
\setlength{\tabcolsep}{4pt}
\caption{Switch operator $\times$ switch epoch grid on \textsc{FrankaCabinet}. Values are means over eight seeds. No operator uniformly dominates across timing and metrics.}
\label{tab:app-switch-grid}
\begin{tabular}{lccccccc}
\toprule
Switch epoch & Regime & Operator & $n$ & Final $\uparrow$ & Max $\uparrow$ & Critic peak $\downarrow$ & $|\Delta \bar r|$ \\
\midrule
50 & immature & hard switch & 8 & \textbf{0.1200} & 0.1845 & \textbf{0.5430} & 0.2073 \\
50 & immature & PBRS & 8 & 0.0892 & 0.2263 & 1.2868 & 0.2115 \\
50 & immature & critic reset & 8 & 0.0868 & \textbf{0.2959} & 2.4217 & 0.5279 \\
\midrule
500 & sweet spot & hard switch & 8 & 0.0150 & 0.0182 & \textbf{0.4797} & 0.9122 \\
500 & sweet spot & PBRS & 8 & 0.0597 & 0.0607 & 0.4877 & 0.9091 \\
500 & sweet spot & critic reset & 8 & \textbf{0.1421} & \textbf{0.2582} & 3.1131 & 0.9268 \\
\midrule
1000 & later & hard switch & 8 & 0.0776 & \textbf{0.4190} & \textbf{1.1653} & 1.2253 \\
1000 & later & PBRS & 8 & 0.1265 & 0.3103 & 1.2125 & 1.2513 \\
1000 & later & critic reset & 8 & \textbf{0.1466} & 0.3284 & 2.3612 & 1.2370 \\
\bottomrule
\end{tabular}
\end{table}

\begin{figure}[t]
    \centering
    \includegraphics[width=0.98\textwidth]{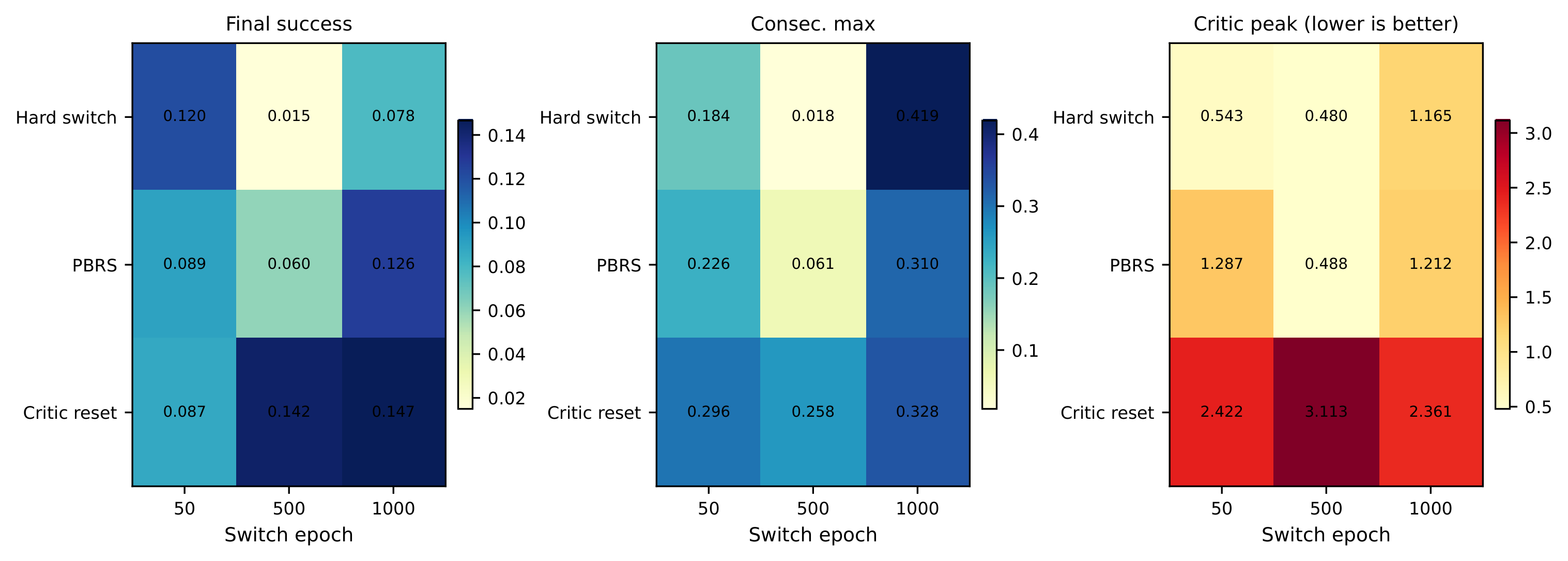}
    \caption{Switching operator heatmaps. Operator performance depends on switch timing and metric. PBRS is not a universal improvement, and critic reset can increase retention while producing high critic instability.}
    \label{fig:app-switch-heatmaps}
\end{figure}

\paragraph{Interpretation.}
The switch-operator results should not be interpreted as evidence for a universal safe-switching mechanism. Hard switching is the practical default in the locked setting because it combines strong peak and tail performance with low critic instability. PBRS can be useful in some regimes but is not robustly dominant. Critic reset can improve terminal retention, but its large critic peaks make it a trade-off rather than a default solution.

\subsection{Selector and Trigger Ablations}
\label{app:selector-controls}

A natural question is whether fixed warm-up can be replaced by an automatic online selector or competence-triggered switch. To avoid keeping the older small-$n$ selector pilot table alongside the revised main-text control framing, we retain only the updated 6x3090 selector-support comparison here.

Table~\ref{tab:app-trigger-results} summarizes the updated selector-support comparison. On \textsc{FrankaCabinet}, the fixed \texttt{wu=50} hard switch remains stronger than the selector variants in the revised support comparison.

\begin{table}[t]
\centering
\small
\setlength{\tabcolsep}{4pt}
\caption{Updated selector support comparison on \textsc{FrankaCabinet}. These rows are \texttt{selector\_support\_6x3090} evidence and replace the older small-$n$ selector ablation table.}
\label{tab:app-trigger-results}
\resizebox{0.98\textwidth}{!}{\begin{tabular}{lcccccc}
\toprule
Method & $n$ & Mean & CI low & CI high & No-switch rate & Mean switches \\
\midrule
\texttt{conservative\_periodic\_selector} & 8 & 0.0508 & 0.0143 & 0.0967 & 0.50 & 0.50 \\
\texttt{direct\_oracle\_or\_direct\_best} & 8 & 0.2150 & 0.0586 & 0.4019 & 1.00 & 0.00 \\
\texttt{fixed\_wu50\_hard} & 8 & 0.3889 & 0.2362 & 0.5226 & 1.00 & 0.00 \\
\texttt{moving\_average\_selector} & 8 & 0.1127 & 0.0680 & 0.1595 & 0.125 & 0.875 \\
\texttt{naive\_last\_selector} & 8 & 0.0919 & 0.0521 & 0.1374 & 0.125 & 0.875 \\
\texttt{one\_shot\_early} & 8 & 0.1252 & 0.0785 & 0.1752 & 1.00 & 0.00 \\
\texttt{one\_shot\_oracle} & 8 & 0.0175 & 0.0068 & 0.0304 & 1.00 & 0.00 \\
\bottomrule
\end{tabular}
}
\end{table}

\begin{table}[tbp]
\centering
\small
\setlength{\tabcolsep}{3.5pt}
\caption{Table 22: Selector failure taxonomy. The appendix records a fuller failure taxonomy than the representative failure patterns visualized in Figure 9.}
\label{tab:app-selector-taxonomy}
\begin{tabular}{lp{0.48\linewidth}}
\toprule
Failure mode & Description \\
\midrule
Never switch
& Selector fails to trigger a reward change despite later-phase evidence. \\
\midrule
Switch without gain
& Selector switches, but the chosen transition does not improve downstream performance. \\
\midrule
No reliable checkpoint
& Verification signal does not satisfy the reliability criterion. \\
\midrule
Over-conservative trigger
& Trigger delays switching until the useful phase window has passed. \\
\midrule
Wrong reward identity
& Selector commits to a candidate whose early local advantage does not translate to downstream utility. \\
\bottomrule
\end{tabular}
\end{table}

\begin{figure}[t]
    \centering
    \includegraphics[width=\linewidth]{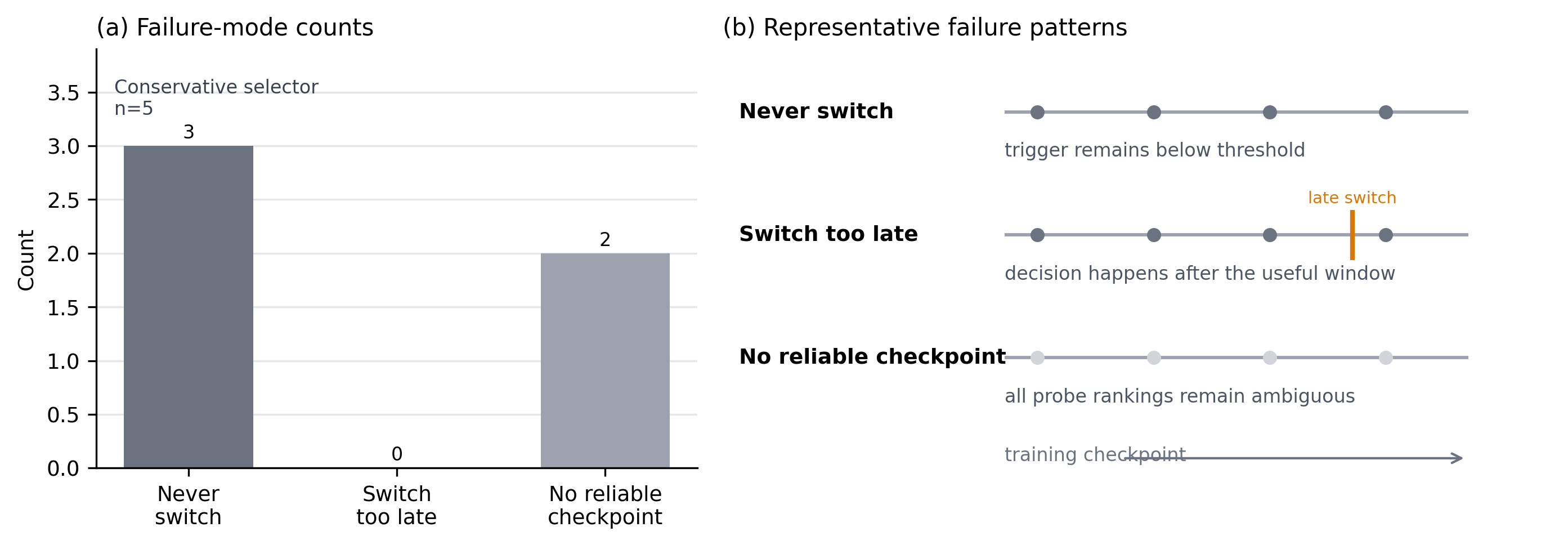}
    \caption{Selector and trigger diagnostics. The figure visualizes the dominant reactive-selector failure patterns: never switching, switching too late, or failing to find a reliable checkpoint.}
    \label{fig:app-selector-timing}
\end{figure}

Figure~\ref{fig:app-selector-timing} visualizes the dominant selector failure patterns, while Table~\ref{tab:app-selector-taxonomy} records the fuller failure taxonomy used in the appendix diagnostics.

\paragraph{Blocked BallBalance selector rows.}
Two \textsc{BallBalance} selector rows remain documented as blocked appendix entries: the moving-average selector and a legacy globally tuned critic-loss threshold selector. They are not part of the locked main evidence and do not affect the boundary conclusion, since \textsc{BallBalance} peak success saturates for all main methods.

\paragraph{Interpretation.}
The selector ablations support the paper's main positioning. \textsc{RHyVE} is not strongest as an increasingly reactive scheduler. Its value lies in competence-aware verification and stable phase-aware deployment. In the sparse small-candidate setting, a simple fixed schedule can be more reliable than reactive selectors that depend on noisy early verification signals.
\section{Generalization, Stress Tests, and Scope}
\label{app:generalization-scope}

This appendix reports evidence that delimits the scope of \textsc{RHyVE}. The following subsections cover LLM-generated candidate families, cross-task breadth, candidate-pool scaling, and support-only or negative evidence retained for transparency.

\subsection{LLM-Generated Candidate Families}
\label{app:llm-source}

The main experiments use small structured reward-hypothesis families so that verification and deployment effects can be cleanly analyzed. To test whether the observed phase-dependence is limited to hand-crafted reward families, we also evaluate LLM-generated candidate pools. For each task, candidate rewards are generated using multiple prompt variants, temperatures, and replicates. For analysis, we distinguish a hand-crafted reference family and two updated LLM source families (lightly curated and minimally filtered). The hand-crafted reference family is reported in the locked main comparison and in the held-out detail table, while Table 24 focuses on the updated LLM source families.

The purpose of this experiment is not to prove that a fixed warm-up schedule is universally optimal for LLM-generated reward pools. Rather, it tests whether phase-dependent winner changes also arise when candidate rewards are generated by an upstream language model.

\begin{table}[t]
\centering
\small
\setlength{\tabcolsep}{4pt}
\caption{LLM candidate-generation statistics. The table summarizes generated, parsed, valid, and executable candidates. These statistics are used to characterize candidate-source diversity, not to support main performance claims.}
\label{tab:app-llm-candidate-stats}
\begin{tabular}{lcccccc}
\toprule
Task & Generated & Parsed & Executable rate & Eval family size & Pool size & Semantic diversity \\
\midrule
\textsc{FrankaCabinet} & 105 & 105 & 0.905 & 3 & 10 & 10.76--34.41 \\
\textsc{BallBalance} & 108 & 108 & 0.787 & 3 & 10 & 11.02--12.10 \\
\textsc{Cartpole} & 108 & 108 & 0.972 & 3 & 10 & 6.88--13.47 \\
\bottomrule
\end{tabular}
\end{table}

Table~\ref{tab:app-source-clean} reports the updated reduced 6x3090 LLM source-generalization summary, with deployable rows separated from oracle references by an explicit status field. Table~\ref{tab:app-source-partial} records the corresponding held-out schedule-selection details for the deployable source families.

\begin{table}[t]
\centering
\small
\setlength{\tabcolsep}{4pt}
\caption{Updated reduced 6x3090 LLM source-generalization summary. Deployable rows and oracle references are separated explicitly so that the appendix does not preserve the older pilot-style completed-only summary as if it were current main evidence. Oracle rows are non-deployable references and are not guaranteed to upper-bound deployable rows when seed sets or comparison protocols differ.}
\label{tab:app-source-clean}
\resizebox{0.98\textwidth}{!}{\begin{tabular}{llcccc}
\toprule
Candidate source & Method & $n$ & Max / \texttt{consec\_max} $\uparrow$ & Flip rate & Status \\
\midrule
\texttt{llm\_lightly\_curated} & Direct & 8 & 0.145 & 0.875 & deployable \\
\texttt{llm\_lightly\_curated} & \texttt{wu=50} hard & 8 & 0.207 & 0.875 & deployable \\
\texttt{llm\_lightly\_curated} & Held-out selected & 5 & 0.298 & -- & deployable \\
\cmidrule(lr){2-6}
\texttt{llm\_lightly\_curated} & Best adaptive (oracle reference) & 3 & 0.580 & 0.875 & oracle \\
\midrule
\texttt{llm\_minimally\_filtered} & Direct & 8 & 0.211 & 0.500 & deployable \\
\texttt{llm\_minimally\_filtered} & \texttt{wu=50} hard & 8 & 0.639 & 0.500 & deployable \\
\texttt{llm\_minimally\_filtered} & Held-out selected & 5 & 0.800 & -- & deployable \\
\cmidrule(lr){2-6}
\texttt{llm\_minimally\_filtered} & Best adaptive (oracle reference) & 2 & 0.369 & 0.500 & oracle \\
\bottomrule
\end{tabular}
}
\end{table}

\begin{table}[t]
\centering
\small
\setlength{\tabcolsep}{3.2pt}
\caption{Held-out schedule-selection details for the reduced 6x3090 source families. Non-deployable oracle references are recorded explicitly but kept separate from deployable held-out rows in the main text.}
\label{tab:app-source-partial}
\resizebox{\columnwidth}{!}{\begin{tabular}{lllccl}
\toprule
Candidate source & Dev seeds & Test seeds & Selected schedule & Evaluated method & Notes \\
\midrule
\texttt{structured\_handcrafted\_K3} & 42,123,456 & 789,999,1234,2025,3031 & \texttt{fixed\_wu50\_hard} & \texttt{fixed\_wu50\_hard} & dev\_prefers\_fixed\_wu50 \\
\texttt{structured\_handcrafted\_K3} & 42,123,456 & 789,999,1234,2025,3031 & \texttt{oracle\_upper\_bound} & \texttt{best\_adaptive\_upper\_bound} & non-deployable reference \\
\texttt{llm\_lightly\_curated} & 42,123,456 & 789,999,1234,2025,3031 & \texttt{direct\_oracle\_or\_direct\_best} & \texttt{direct\_oracle\_or\_direct\_best} & dev\_prefers\_direct \\
\texttt{llm\_lightly\_curated} & 42,123,456 & 789,999,1234,2025,3031 & \texttt{oracle\_upper\_bound} & \texttt{best\_adaptive\_upper\_bound} & non-deployable reference \\
\texttt{llm\_minimally\_filtered} & 42,123,456 & 789,999,1234,2025,3031 & \texttt{fixed\_wu50\_hard} & \texttt{fixed\_wu50\_hard} & dev\_prefers\_fixed\_wu50 \\
\texttt{llm\_minimally\_filtered} & 42,123,456 & 789,999,1234,2025,3031 & \texttt{oracle\_upper\_bound} & \texttt{best\_adaptive\_upper\_bound} & non-deployable reference \\
\bottomrule
\end{tabular}
}
\end{table}

\begin{figure}[t]
    \centering
    \begin{minipage}[t]{0.49\linewidth}
        \centering
        \includegraphics[width=\linewidth]{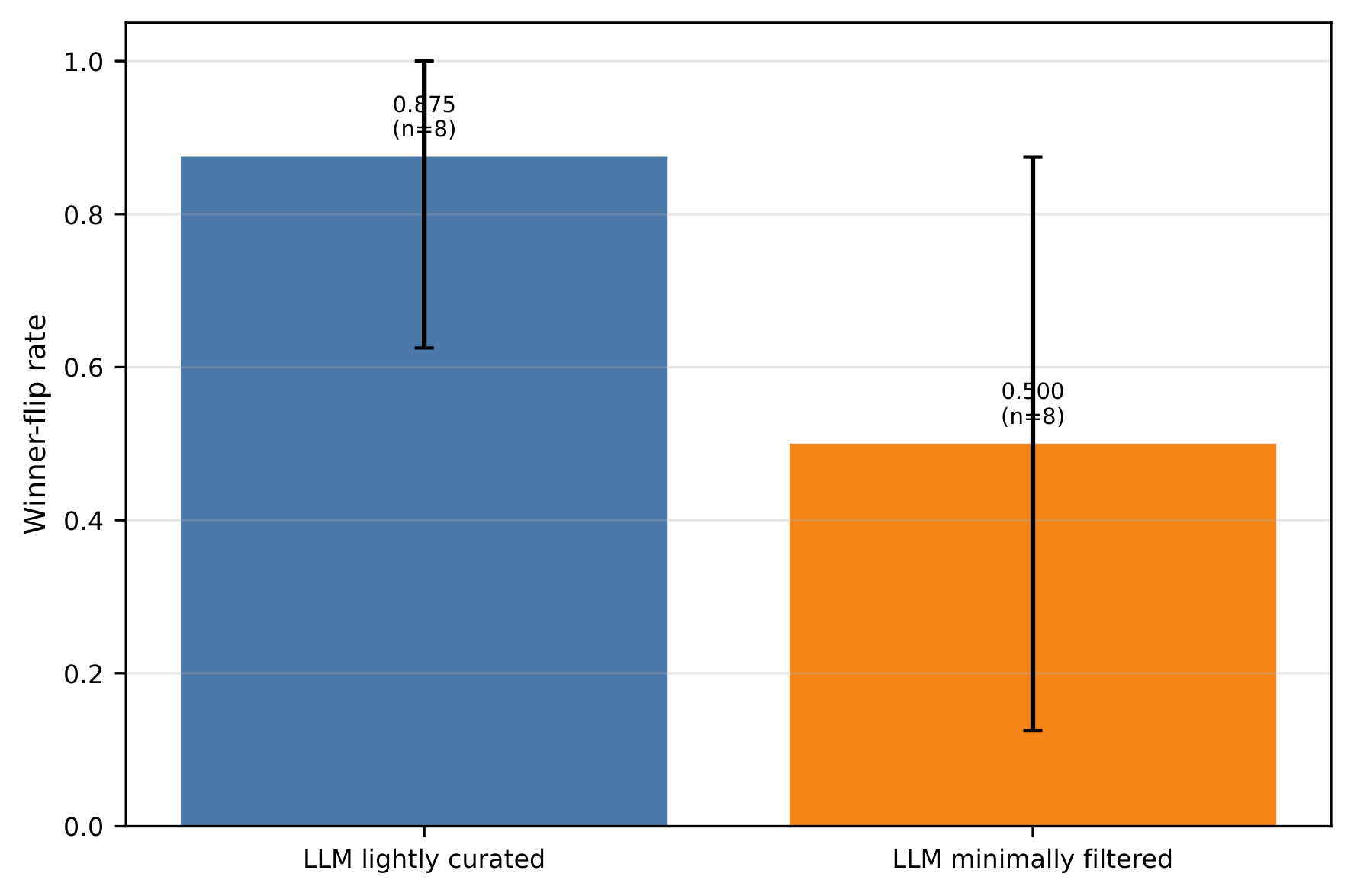}
    \end{minipage}\hfill
    \begin{minipage}[t]{0.49\linewidth}
        \centering
        \includegraphics[width=\linewidth]{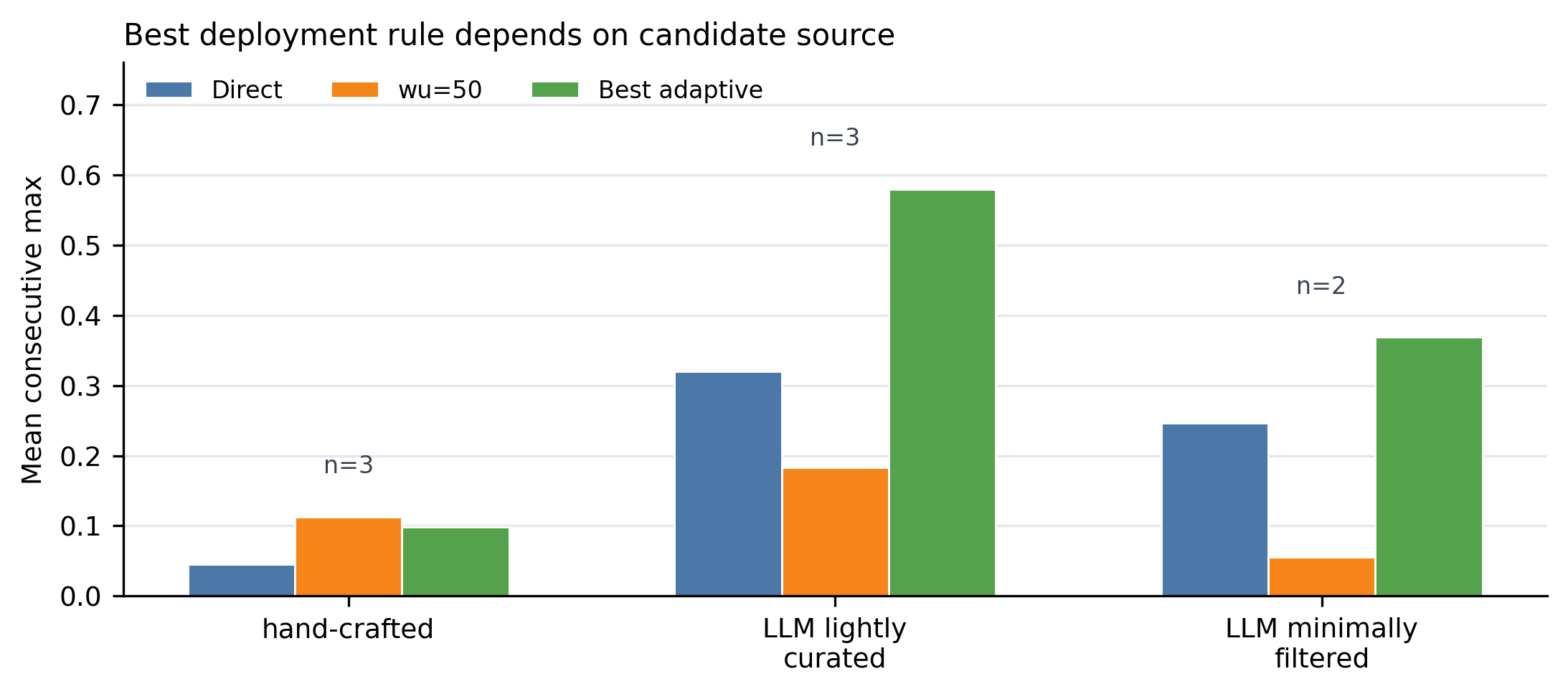}
    \end{minipage}
    \caption{Left: Winner-flip rates across candidate sources. LLM-generated \textsc{FrankaCabinet} reward families exhibit phase-dependent winner changes, while dense or ceiling-prone tasks show little variation. Right: Method comparison across reward-candidate sources. The best deployment rule can vary across generated pools, so this experiment is used as phenomenon evidence rather than a fixed-schedule superiority claim.}
    \label{fig:app-source-flip}
    \label{fig:app-source-methods}
\end{figure}

\paragraph{Takeaway.}
The source-generalization experiments support the hypothesis that phase-dependent reward utility is not an artifact of a hand-designed reward family. However, they also show why \textsc{RHyVE} should be framed as a verification protocol rather than a fixed universal schedule: different generated pools can favor different deployment rules.

\subsection{Cross-Task Breadth}
\label{app:cross-task}

Table~\ref{tab:app-cross-task} summarizes the cross-task phenomenon scorecard. The goal of this table is not benchmark dominance. Instead, it records whether the main phenomena appear across tasks and where the method's assumptions break down, including the optional \textsc{FrankaCubeStack} all-failure boundary.

\begin{table}[t]
\centering
\small
\setlength{\tabcolsep}{4pt}
\caption{Cross-task phenomenon scorecard. The strongest practical evidence is on \textsc{FrankaCabinet}; other tasks are used to establish boundary conditions and phenomenon breadth.}
\label{tab:app-cross-task}
\resizebox{0.98\textwidth}{!}{\begin{tabular}{lccccp{0.28\linewidth}}
\toprule
Task & Competence threshold & Winner flip & Warm-up helps & Evidence & Interpretation \\
\midrule
\textsc{FrankaCabinet} & task-dependent & observed & yes & \texttt{locked\_main\_existing}/\texttt{reduced\_main\_6x3090} & main sparse phase-sensitive evidence \\
\textsc{BallBalance} & not limiting & not central & not meaningful (ceiling) & \texttt{historical\_legacy} & dense boundary control \\
\textsc{CartPole} & weak & fragile & not reliable & \texttt{historical\_legacy} & appendix breadth boundary \\
\textsc{Ant} & weak & fragile & not reliable & \texttt{historical\_legacy} & appendix breadth boundary \\
\textsc{FrankaCubeStack} & not meaningful & not meaningful & not meaningful & \texttt{optional\_extra\_task\_pilot\_6x3090} & all-failure boundary; not used for method ranking \\
\bottomrule
\end{tabular}
}
\end{table}

The cross-task results support a task-aware interpretation. \textsc{RHyVE} is most useful when the task is sparse or phase-sensitive, such that early and late rewards serve different roles. When a dense reward already provides sufficient guidance, or when reward rankings do not exhibit meaningful phase changes, warm-up may be unnecessary.

\subsection{Candidate-Pool Scaling Stress Test}
\label{app:pool-scaling}

The main method is designed for small candidate sets. To test the boundary of this regime, we perform candidate-pool scaling stress tests with $K\in\{3,5,10\}$ on \textsc{FrankaCabinet} and \textsc{Cartpole}. These results are marked as \texttt{stress\_tests} and are not used for main performance claims.

\begin{table}[t]
\centering
\small
\setlength{\tabcolsep}{4pt}
\caption{Candidate-pool scaling stress test. Larger pools increase verification cost and can reduce top-1 reliability. This supports the small-candidate local-verification scope of \textsc{RHyVE}.}
\label{tab:app-pool-scaling}
\resizebox{0.98\textwidth}{!}{\begin{tabular}{lccccc}
\toprule
Task & $K$ & Top-1 hit rate & Hit@3 & Rank corr. & Overhead (s) \\
\midrule
\textsc{FrankaCabinet} & 3 & 1.00 & 1.00 & 0.833 & 1732 \\
\textsc{FrankaCabinet} & 5 & 1.00 & 1.00 & 0.950 & 2909 \\
\textsc{FrankaCabinet} & 10 & 0.50 & 1.00 & 0.911 & 5850 \\
\midrule
\textsc{CartPole} & 3 & 0.50 & 1.00 & 0.500 & 60 \\
\textsc{CartPole} & 5 & 0.00 & 0.50 & 0.700 & 100 \\
\textsc{CartPole} & 10 & 0.00 & 0.50 & 0.600 & 201 \\
\bottomrule
\end{tabular}
}
\end{table}

\begin{figure}[t]
    \centering
    \begin{minipage}[t]{0.49\linewidth}
        \centering
        \includegraphics[width=\linewidth]{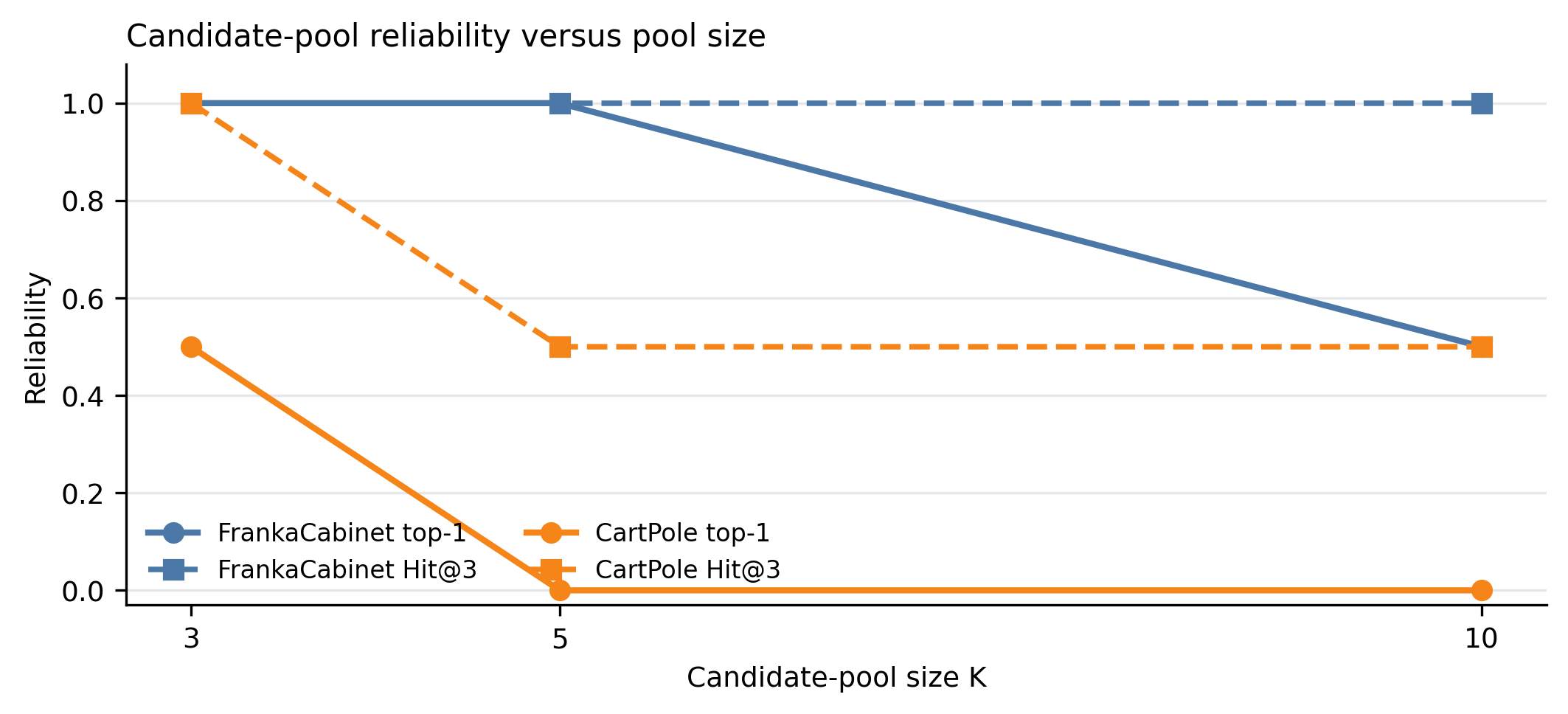}
    \end{minipage}\hfill
    \begin{minipage}[t]{0.49\linewidth}
        \centering
        \includegraphics[width=\linewidth]{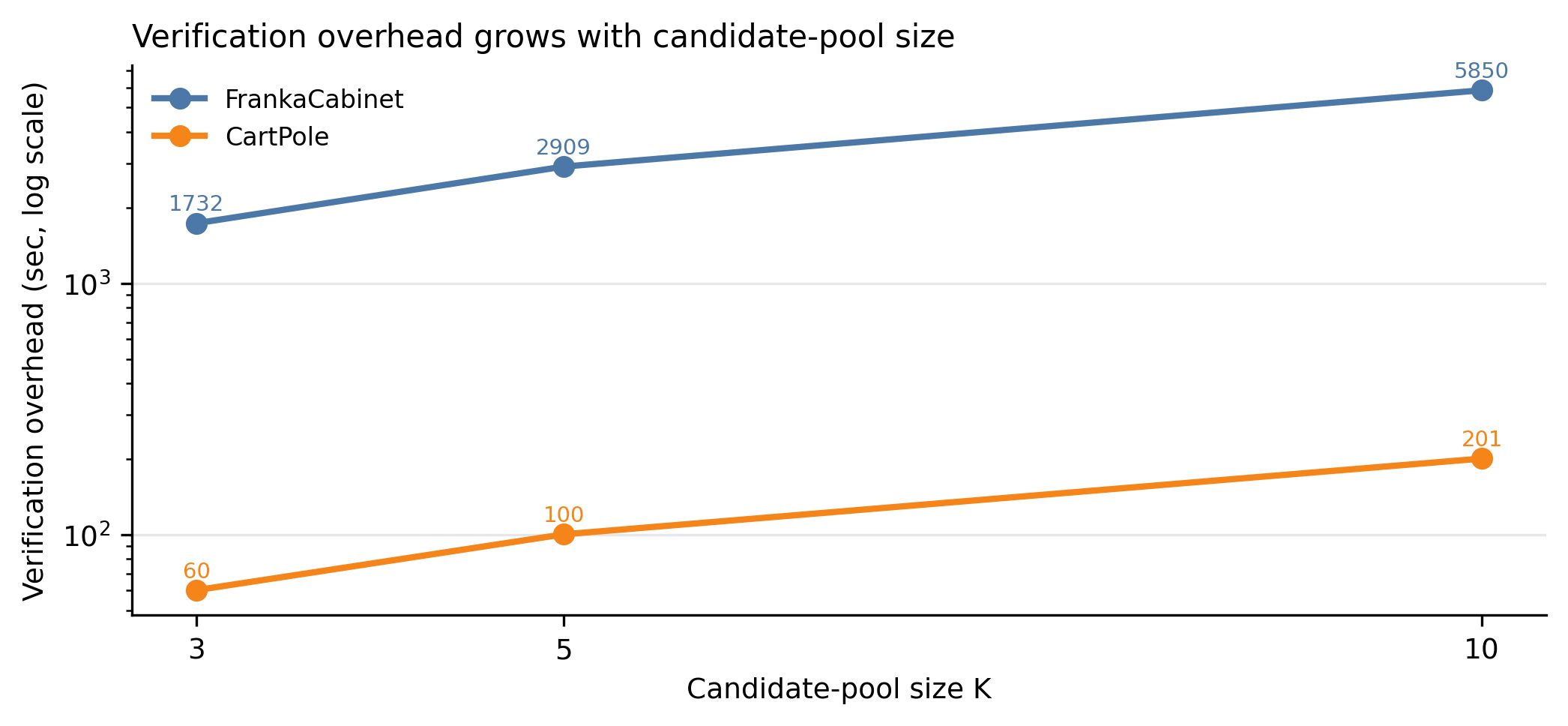}
    \end{minipage}
    \caption{Left: Candidate-pool reliability versus pool size. Top-1 reliability weakens as the candidate set grows, especially on the lightweight stress task. Right: Verification overhead grows with candidate-pool size. This motivates using \textsc{RHyVE} as a small-candidate local-verification protocol rather than exhaustive large-pool search.}
    \label{fig:app-pool-reliability}
    \label{fig:app-pool-compute}
\end{figure}

\paragraph{Takeaway.}
The pool-scaling experiment narrows the paper's claim. \textsc{RHyVE} is currently best supported for small candidate sets. Extending it to larger pools will likely require coarse-to-fine search, uncertainty-aware ranking, clustering, or more efficient candidate pruning.

\subsection{Additional Limitations, Blocked Runs, and Negative Evidence}
\label{app:negative-support}

We include negative and blocked evidence for transparency. These results are not used for main quantitative claims, but they clarify the boundary of the current method.

\begin{table}[t]
\centering
\small
\setlength{\tabcolsep}{4pt}
\caption{Support-only and negative evidence summary. These results are documented to clarify scope and prevent overclaiming.}
\label{tab:app-negative-summary}
\begin{tabular}{p{0.25\linewidth}p{0.25\linewidth}p{0.40\linewidth}}
\toprule
Evidence block & Status & Interpretation \\
\midrule
BallBalance blocked selectors
& documented, not rerun
& Two selector rows remain blocked; they are appendix-level support baselines and do not affect the locked BallBalance boundary conclusion. \\
\midrule
Trigger-based switching
& support only
& Competence-triggered switching does not replace the fixed \texttt{wu=50} schedule on \textsc{FrankaCabinet}. \\
\midrule
Reactive selectors
& support only
& Selectors can fail by never switching, switching too late, or selecting unstable winners. \\
\midrule
Candidate-pool scaling
& stress test
& Larger pools weaken top-1 reliability and increase compute cost. \\
\midrule
Cartpole and Ant breadth
& boundary / negative
& These tasks show that phase dependence and warm-up gains are task-dependent rather than universal. \\
\midrule
Historical singleton runs
& legacy only
& Retained for traceability but excluded from main and support claims. \\
\bottomrule
\end{tabular}
\end{table}

\paragraph{Scope of the current evidence.}
The most defensible interpretation of the full experimental suite is as follows. \textsc{RHyVE} identifies and operationalizes a real deployment issue for LLM-generated rewards: reward hypotheses may become reliable and useful only after sufficient learner competence. On sparse phase-sensitive tasks, this can make a simple phase-aware schedule stronger than direct or one-shot deployment. However, the method is not a universal adaptive scheduler, not a large-pool reward-search algorithm, and not a guarantee that warm-up improves every task.

\paragraph{Recommended use.}
In its current form, \textsc{RHyVE} is most appropriate when:
\begin{enumerate}
    \item a small set of plausible reward hypotheses is available;
    \item the task is sparse or phase-sensitive;
    \item early reward comparisons are suspected to be unreliable;
    \item the practitioner can afford sparse shared-checkpoint fork verification;
    \item the deployment goal is to choose a stable phase-aware schedule rather than continuously chase a reactive selector.
\end{enumerate}

When the oracle-like reward is already dense, or when the candidate set is too large for reliable local ranking, \textsc{RHyVE} should be used as a diagnostic rather than as an automatic deployment rule.

\subsection{FrankaCubeStack Optional Pilot}
\label{app:frankacubestack-pilot}

The reduced optional-scope \textsc{FrankaCubeStack} pilot is retained only as all-failure boundary evidence. It is not used for method ranking, and it does not change the main task-aware interpretation of the paper.

\begin{table}[t]
\centering
\small
\setlength{\tabcolsep}{4pt}
\caption{\textsc{FrankaCubeStack} optional pilot boundary table. All rows are \texttt{optional\_extra\_task\_pilot\_6x3090} evidence and are retained only to document the all-failure boundary.}
\label{tab:app-frankacubestack-boundary}
\resizebox{0.98\textwidth}{!}{\begin{tabular}{lcccccccc}
\toprule
Method & $n$ & Max / \texttt{consec\_max} $\uparrow$ & Final $\uparrow$ & Last-5 mean $\uparrow$ & Full AUC $\uparrow$ & Tail AUC $\uparrow$ & Collapse rate & Status \\
\midrule
\texttt{direct\_late\_task\_faithful} & 3 & 0.0000 & 0.0000 & 0.0000 & 0.0000 & 0.0000 & 0.0 & \texttt{optional\_extra\_task\_pilot\_6x3090} \\
\texttt{fixed\_wu100\_hard} & 3 & 0.0028 & 0.0000 & 0.0000 & 0.000084 & 0.000005 & 0.0 & \texttt{optional\_extra\_task\_pilot\_6x3090} \\
\texttt{fixed\_wu50\_hard} & 3 & 0.0001 & 0.0000 & 0.0000 & 0.0000003 & 0.0000 & 0.0 & \texttt{optional\_extra\_task\_pilot\_6x3090} \\
\bottomrule
\end{tabular}
}
\end{table}

\begin{figure}[t]
    \centering
    \includegraphics[width=\linewidth]{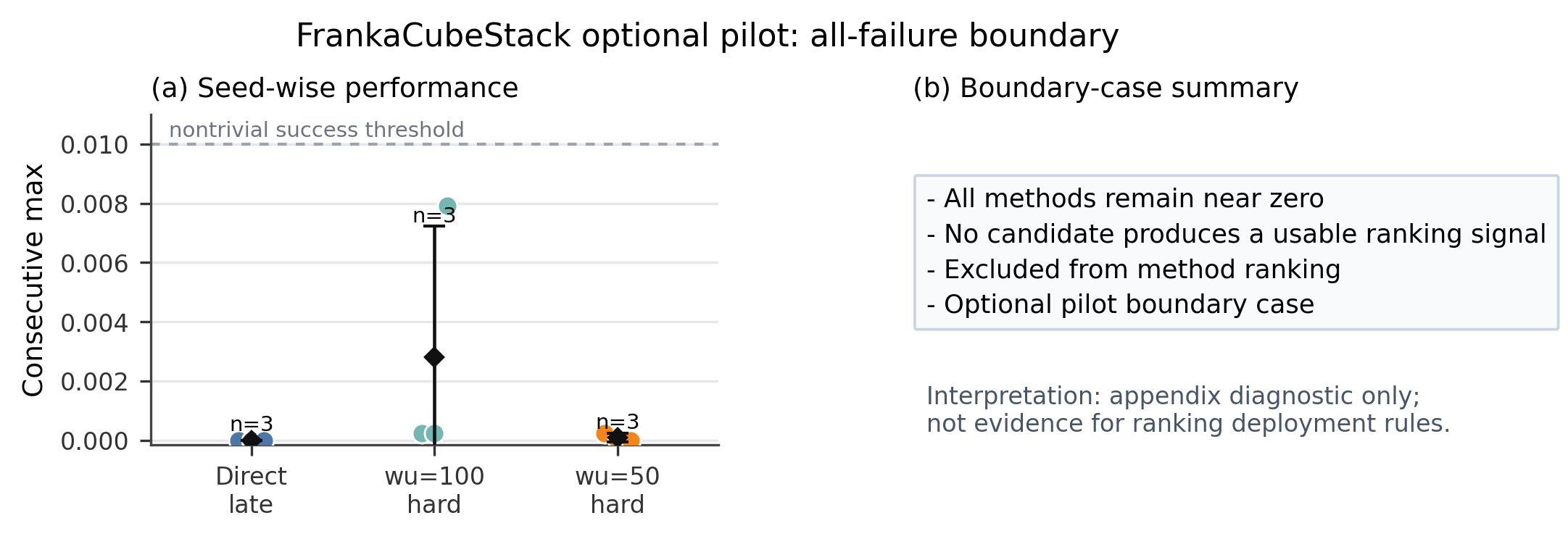}
    \caption{\textsc{FrankaCubeStack} optional pilot: all-failure boundary. All compared methods remain effectively flat under the reduced optional-scope budget, so this pilot is retained only as boundary evidence and not used for method ranking.}
    \label{fig:app-frankacubestack-boundary}
\end{figure}






\end{document}